\documentclass{article}
\usepackage[T1]{fontenc}
\usepackage{microtype}
\usepackage{graphicx}
\usepackage{booktabs}
\usepackage{amsmath,amssymb}
\usepackage{array,multirow}
\usepackage[table]{xcolor}
\PassOptionsToPackage{hyphens}{url}
\usepackage{hyperref}
\usepackage[accepted]{mlsys2025}
\definecolor{sqshade}{RGB}{239,244,249}
\definecolor{draftgray}{RGB}{82,89,98}
\definecolor{boxgray}{RGB}{150,156,164}
\definecolor{boxfill}{RGB}{248,249,251}
\newif\ifillustrativeplots
\illustrativeplotstrue
\newcommand{\sq}{VR-SparseQuant}
\newcommand{\sqshort}{VR-SQ}
\newcommand{\pending}{\textcolor{draftgray}{\textemdash}}
\RequirePackage{etoolbox}
\AtBeginEnvironment{figure}{\setlength{\parskip}{0pt}}
\AtBeginEnvironment{figure*}{\setlength{\parskip}{0pt}}
\AtBeginEnvironment{table}{\setlength{\parskip}{0pt}}
\AtBeginEnvironment{table*}{\setlength{\parskip}{0pt}}
\patchcmd{\section}{-0.12in}{-0.08in}{}{}
\patchcmd{\subsection}{-0.10in}{-0.07in}{}{}
\patchcmd{\paragraph}{1.5ex plus}{0.8ex plus}{}{}
\makeatletter
\def\sq@figuretype{figure}
\patchcmd{\@makecaption}{\vskip 10pt}{%
  \ifx\@captype\sq@figuretype
    \vskip 2.5pt\relax
  \else
    \vskip 4pt\relax
  \fi
}{}{\PackageError{SparseQuant}{Caption spacing could not be adjusted}%
  {Check the MLSys caption definition.}}
\makeatother
\newcommand{\sqtablebreak}[1]{\noalign{\vskip1.2pt
  \hbox to#1{\color{gray!65}\xleaders\hbox{\rule{3pt}{0.35pt}\hskip3pt}\hfill}
  \vskip1.2pt}}
\newcommand{\sqablationstyle}{\footnotesize
  \setlength{\tabcolsep}{0.5pt}\renewcommand{\arraystretch}{1.0}}

\newcommand{\tabnote}[1]{\par\smallskip{\footnotesize\raggedright #1\par}}

\mlsystitlerunning{VR-SparseQuant: Accurate FP4 Quantization with 2:4 Sparsity on Vera Rubin}
\makeatletter
\newenvironment{firstpageteaser}{%
  \par\begin{minipage}{\textwidth}\def\@captype{figure}\centering
}{\end{minipage}\par}
\makeatother

\begin{document}
\twocolumn[
\mlsystitle{VR-SparseQuant: Accurate FP4 Quantization\texorpdfstring{\\}{ }with 2:4 Sparsity on Vera Rubin}
% \begin{mlsysauthorlist}
% \mlsysauthor{Anonymous Authors}{anon}
% \end{mlsysauthorlist}
% \mlsysaffiliation{anon}{Anonymous Institution}

\begin{mlsysauthorlist}
\mlsysauthor{Samir Khaki}{to}
\mlsysauthor{Wei-Chen Wang}{to}
\mlsysauthor{Song Han}{goo}
\mlsysauthor{Ryan Hanrui Wang}{to}
\end{mlsysauthorlist}

\mlsysaffiliation{to}{Nebius Applied AI, San Francisco, CA, USA}
\mlsysaffiliation{goo}{Massachusetts Institute of Technology, Cambridge, MA, USA}

\mlsyscorrespondingauthor{Ryan Hanrui Wang}{wang@nebius.com}

\mlsyskeywords{quantization, sparse tensor cores, language models, vision-language models, inference}
\vskip 0.06in
\begin{firstpageteaser}
\includegraphics[width=\linewidth]{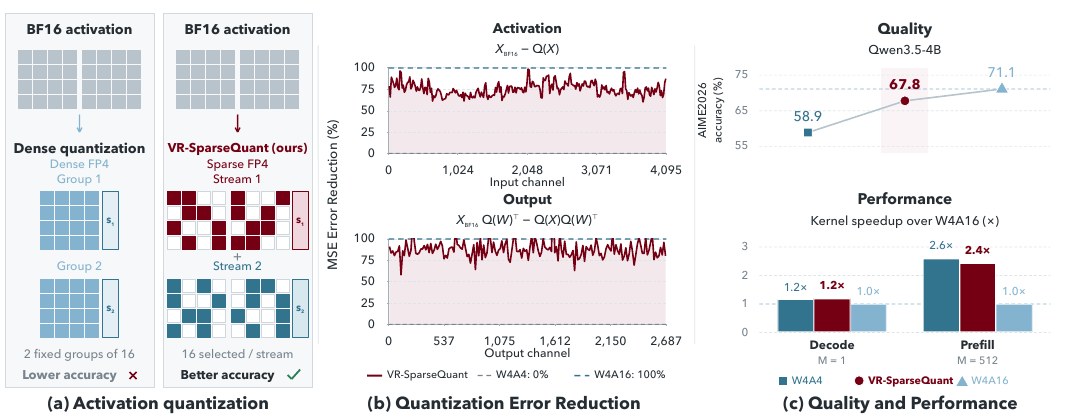}
\caption{\textbf{Structured sparsity for accurate and efficient FP4 inference.}
\textbf{(a)} Two 2:4 sparse FP4 streams approximate a dense activation
block while retaining dense FP4's value budget. Overlapping positions
receive a second contribution that reduces their reconstruction error.
\textbf{(b)} Plots show the reduction in activation and layer-output mean
squared error (MSE) achieved by \sq{}.
\textbf{(c)} \sq{} improves Qwen3.5-4B AIME2026 accuracy over
W4A4; Rubin GEMMs retain similar speed to dense FP4 and outperform W4A16.}
\label{fig:teaser}
\end{firstpageteaser}

\vskip 0.08in
\begin{abstract}
In recent years, GPUs have incorporated increasingly specialized support
for low-precision computation, from Hopper's FP8 to Blackwell's FP4 Tensor
Cores. Precision formats have evolved alongside
this hardware, from MXFP4 and NVFP4 on Blackwell to E2M1 FP4 with direct
UE5M3 block scaling on Vera Rubin, which we term as \textbf{VRFP4}.
These advances help serve larger language models at lower cost,
but four-bit activation quantization can substantially degrade model
quality. We introduce \sq{}, which uses Vera Rubin's elementwise 2:4
sparsity to recover accuracy while retaining native FP4 execution.
By allowing two sparse streams to overlap, we use the same FP4 value
budget as dense quantization to refine activations with larger errors.
Offline scale-ratio calibration avoids runtime search, while a
shared-weight GEMM accumulates both streams efficiently.
On Nemotron-3-Nano-30B-A3B, \sq{} reduces WikiText-2 perplexity from
\textbf{6.85} to \textbf{5.57} versus dense RTN W4A4. Including activation
preparation, our native Rubin kernels for VRFP4 achieve up to
$\boldsymbol{3.1\times}$
GEMM and $\boldsymbol{2.7\times}$ MoE speedup over BF16.
To the best of our knowledge, this is the first study of adaptive
activation quantization using Vera Rubin's elementwise 2:4 structured
sparse FP4 instructions.

\end{abstract}
]
\printAffiliationsAndNotice{}
\hypersetup{pdfsubject={MLSys 2027 manuscript}}
\raggedbottom
\setlength{\parskip}{2.5pt plus 0.5pt}
\setlength{\abovedisplayskip}{4pt plus 1pt minus 1pt}
\setlength{\belowdisplayskip}{4pt plus 1pt minus 1pt}
\setlength{\belowdisplayshortskip}{3pt plus 1pt minus 1pt}
\section{Introduction}
\label{sec:intro}
Recent advances in large language model (LLM) inference have made
low-bit execution practical for interactive generation and large-batch
serving~\cite{awq2024,qserve2024}. These workloads occupy different points
on the latency--throughput curve, with bottlenecks that change as weights
are reused. At small decode batches, reading weights can dominate linear
layers; larger batches and prompt processing reuse them across more
tokens, making matrix-multiplication throughput increasingly important.
Low-bit inference must therefore reduce both weight traffic and
arithmetic cost to serve these operating regimes.

Weight-only quantization lowers memory footprint and traffic while retaining
higher-precision activations~\cite{awq2024}. Native FP4 also lowers compute
cost, but requires four-bit activations encoded online, unlike weights
quantized and packed offline. This encoding must preserve quality with
little work and fit native instructions. Preparing models for NVIDIA's
Vera Rubin platform therefore requires designing quantization around its
arithmetic, scales, and sparse layouts~\cite{nvidiarubinplatform2026,nvidiaptx2026}.

Dense FP4 gives every activation one four-bit code and a shared block
scale. Fixed slots simplify online packing but allocate values uniformly
despite unequal errors. When large values set the scale, smaller values
may be poorly resolved; smoothing and adaptive scales improve the
fit~\cite{smoothquant2023,fouroversix2025}. Some positions still benefit
more from a second contribution correcting rounding error than others.
Existing methods add residual channels or sparse--dense
branches~\cite{arcquant2026,sharq2026,rolloutresq2026}. Reallocating a fixed
value budget instead requires choosing which positions to cover and
correct while constructing native operands online.

Structured sparsity makes these choices through local position metadata
while keeping value count and operand format fixed. Blackwell's compact
sparse FP4 path selects adjacent pairs; Vera Rubin selects two individual
positions per quartet~\cite{nvidiaptx87,nvidiaptx2026}. Neither variable-length
records nor input-dependent weight permutations are needed.

We introduce \sq{} (\sqshort{}), which represents an activation as the sum
of two structured 2:4 sparse matrices with different scales to better
separate value ranges and improve quantization accuracy. We refer to each
matrix as a \emph{stream}, which retains the input's logical shape but
stores only two FP4 values per quartet, together with scales and position
metadata.
Together, they match dense FP4's value count and nonzero arithmetic,
but their sparsity patterns can overlap: the fine stream can represent an
uncovered value or correct the coarse stream's approximation. Overlap
leaves another position uncovered, so it is useful when correction
removes more error than the omission introduces
(Figure~\ref{fig:teaser}a).

Scales and positions are coupled, so a joint search repeatedly encodes
each block. After smoothing and weight quantization, we instead calibrate
the scale ratio offline. It ties the fine scale to the coarse scale while
both track each block's range. Two local selections then assign coarse
values and fine residual corrections without searching scale pairs online.

A preparation kernel constructs both streams together. A single sparse
GEMM shares staged weight tiles across both streams and uses one FP32
accumulator, avoiding duplicate global-memory weight-tile loads,
materialized intermediate outputs, and a combining pass. The kernels
apply Rubin's UE5M3 block scales directly to E2M1 values, a format we call
\textbf{VRFP4}.

We evaluate dense and mixture-of-experts language models and vision-language
models on language modeling, reasoning, agentic tasks, and visual
understanding. GPT-OSS~\cite{gptoss2025} and
MiniMax~\cite{minimaxm31config2026} test adaptation from released MXFP4
and NVFP4 checkpoints without original BF16 weights. Coverage studies
progressively quantize more projection families; ablations examine
smoothing, scales, and assignment.
Including online activation encoding, native GEMM and MoE on Vera Rubin
achieve $1.7$--$3.1\times$ and $1.4$--$2.7\times$ speedup over BF16,
respectively; dense FP4 comparisons quantify the additional latency of
this allocation.

To the best of our knowledge, this is the first study to design and
evaluate adaptive activation quantization using Vera Rubin's compact
elementwise sparse FP4 instructions. It connects a representation that
improves model quality with kernels that execute it natively on Rubin.
The broader contribution is a practical use of structured sparsity to
allocate precision: sparse position metadata lets each input determine
where a fixed budget of FP4 values is spent.

\section{Preliminaries: FP4 on Vera Rubin}
\label{sec:background}
For a linear layer $Y=XW^\top$ with $M$ rows, $K$ input channels,
and $N$ output channels, the representation depends on which positions
an operand may store and which values share a scale.

\subsection{Elementwise Sparse FP4}
\label{sec:two-stream-motivation}
A 2:4 sparse operand stores two values and their positions per quartet
along $K$. Sparse Tensor Cores multiply only these entries, halving
nonzero arithmetic relative to a dense product.

\begin{figure}[!tb]
\centering
\includegraphics[width=\columnwidth]{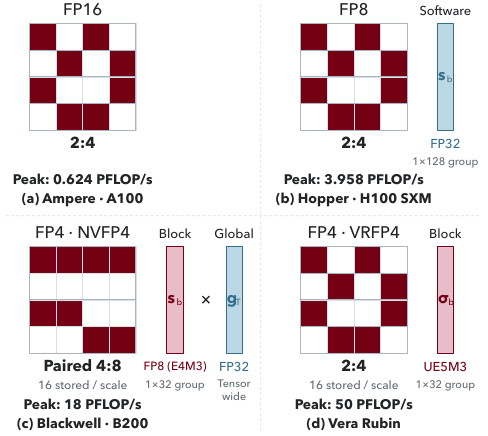}
\caption{\textbf{Rubin brings elementwise 2:4 sparsity to the compact FP4 path.}
Grids show 16 logical positions along $K$. Blackwell's compact FP4
path selects adjacent pairs; Rubin's elementwise path selects individual
values within each quartet. Bars show scaling: the Hopper example uses
software group scales, Blackwell uses FP8 block and FP32 global scales,
and our Rubin path uses direct UE5M3
scales~\cite{nvidiaptx87,nvidiaptx2026,nvidiafp8blockwise}.
Throughput labels are manufacturer per-GPU sparse peaks, not measured
\sq{} performance~\cite{nvidiaa100,nvidiah100,nvidiablackwell,nvidiarubin}.}
\label{fig:gpu-capabilities}
\end{figure}

The allowed positions depend on the instruction format
(Figure~\ref{fig:gpu-capabilities}). Ampere FP16 and Hopper FP8 support
elementwise 2:4 sparsity. Blackwell's compact sparse FP4 path selects
two adjacent pairs from eight positions; Rubin's elementwise FP4 path
selects two individual positions per group of
four~\cite{nvidiaptx87,nvidiaptx2026}. Both retain half the values,
but neither mask family contains the other.

Our selected Rubin path shares one scale across 16 stored values
(32 logical positions) and uses G32 weights.

\subsection{Block Scales and Group Size}
\label{sec:scaling}
Microscaling combines low-bit values with scales shared across small
blocks~\cite{microscaling2023}. E2M1 uses one sign bit, two exponent bits,
and one fraction bit. Group size $G$ governs scale sharing; the scale
format determines range and precision.

NVFP4 combines unsigned E4M3 (UE4M3) scales for 16-value blocks
with an FP32 tensor scale~\cite{nvidianvfp4}:
$q_j=Q_4(t_j/(g_Ts_b))$ and $\hat t_j=g_Ts_bq_j$, where $Q_4$ rounds to E2M1,
$s_b$ is the block scale, and $g_T$ is the tensor scale.
Rubin's UE5M3 retains three fraction bits while widening the exponent
range~\cite{nvidiaptx2026}. We use it for direct block scaling,
$\hat t_j=\sigma_bq_j$, and call the resulting representation VRFP4.
The matrix instruction applies $\sigma_b$ directly, eliminating the
tensor-level $g_Xg_W$ epilogue multiplier. Group size is specified separately.

\paragraph{Comparing scaling schemes at fixed group size.}
Table~\ref{tab:scale-error} compares VRFP4's direct UE5M3 scaling with
NVFP4's bi-level microscaling on
Qwen3.5-4B-Base~\cite{qwen354bbase2026}. Within each G16 or G32 pair,
only the scaling scheme changes. NMSE measures reconstruction error in
the weights, WikiText-2 activations, and resulting outputs of one MLP
down-projection. The two schemes yield nearly identical NMSE and
perplexity in both comparisons, showing that direct block scaling
preserves the quality of bi-level microscaling.
% Source: tables/data/scale_format.json (unrounded values and receipt links).
\begin{table}[!t]
\centering
\caption{\textbf{VRFP4 closely matches NVFP4's reconstruction error.}
Each G16 or G32 pair compares the scaling schemes at fixed group size
under unsmoothed RTN. NMSE uses the weights, sampled activations, and
outputs of one Qwen3.5-4B-Base MLP down-projection; PPL covers all 248
projections.}
\label{tab:scale-error}
\footnotesize
\setlength{\tabcolsep}{1pt}
\renewcommand{\arraystretch}{1.12}
\begin{tabular}{
>{\raggedright\arraybackslash}p{\dimexpr.265\columnwidth-2\tabcolsep\relax}
>{\centering\arraybackslash}p{\dimexpr.065\columnwidth-2\tabcolsep\relax}
>{\centering\arraybackslash}p{\dimexpr.155\columnwidth-2\tabcolsep\relax}
>{\centering\arraybackslash}p{\dimexpr.17\columnwidth-2\tabcolsep\relax}
>{\centering\arraybackslash}p{\dimexpr.155\columnwidth-2\tabcolsep\relax}
>{\centering\arraybackslash}p{\dimexpr.19\columnwidth-2\tabcolsep\relax}}
\toprule
\multirow{2}{*}{\textbf{Format}} & \multirow{2}{*}{$G$}
& \multicolumn{3}{c}{\textbf{NMSE}$\downarrow$}
& \textbf{PPL}$\downarrow$ \\
\cmidrule(lr){3-5}\cmidrule(lr){6-6}
& & Weight & Activation & Output & WikiText-2 \\
\midrule
\rowcolor{gray!10}
\textcolor{draftgray}{BF16} & \textcolor{draftgray}{\textemdash}
& \textcolor{draftgray}{0} & \textcolor{draftgray}{0} & \textcolor{draftgray}{0}
& \textcolor{draftgray}{8.05} \\
\sqtablebreak{\columnwidth}
NVFP4 & 16 & $8.95\mathrm{e}{-3}$ & $8.20\mathrm{e}{-3}$ & $1.51\mathrm{e}{-2}$ & 8.66 \\
VRFP4 & 16 & $8.96\mathrm{e}{-3}$ & $8.21\mathrm{e}{-3}$ & $1.51\mathrm{e}{-2}$ & 8.67 \\
\midrule
NVFP4 & 32 & $1.03\mathrm{e}{-2}$ & $1.13\mathrm{e}{-2}$ & $1.90\mathrm{e}{-2}$ & 8.85 \\
VRFP4 & 32 & $1.03\mathrm{e}{-2}$ & $1.13\mathrm{e}{-2}$ & $1.90\mathrm{e}{-2}$ & 8.86 \\
\bottomrule
\end{tabular}
\end{table}

\section{\sq}
\label{sec:method}
\begin{figure*}[!t]
\centering
\includegraphics[width=\textwidth]{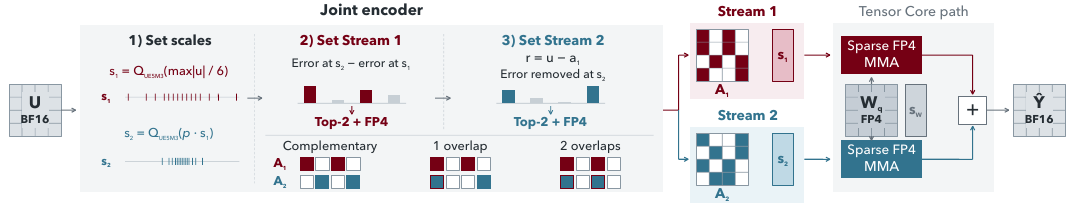}
\vspace{-10pt}
\caption{\textbf{Online allocation within a fixed sparse format.}
The block maximum and a calibrated ratio set two FP4 scales. Greedy
selections allocate coarse values and fine residual corrections within
two fixed-size 2:4 streams. One sparse GEMM reuses quantized weights and
accumulates both products in FP32 before converting the output to BF16.}
\label{fig:method}
\end{figure*}

\sq{} allocates four FP4 values per quartet through two sparse streams;
position metadata adapts this allocation to each input within a fixed
operand layout. Figure~\ref{fig:method} summarizes the workflow: the block
maximum and a calibrated ratio set two grids, two greedy selections
assign values, and one shared-weight GEMM accumulates both products.
% Unmodified author-supplied SparseQuant-Method-v2.pdf, updated 29 October 2026.

\subsection{Two-Stream Quantization}
\label{sec:encoding}
We encode the prepared input $U=\operatorname{BF16}(X\odot b)$, where
$b$ is a positive channel transform fixed during model preparation.
Its inverse is absorbed into the weights (Section~\ref{sec:smoothing}). For a block
$u\in\mathbb{R}^{32}$ of this input, we write $u\approx a_1+a_2$, where each
stream stores two FP4 values per consecutive group of four and one scale
for the block. The scales $S_1$ and $S_2$ define coarse and fine grids.
Across the streams, a logical position can receive zero, one, or two
selected contributions, while each stream retains its fixed 2:4 layout.
A stream's \emph{sparsity pattern} specifies its selected positions;
a selected code can itself be zero. Section~\ref{sec:system} accounts
for the additional metadata and scales.

If the two sparsity patterns are disjoint, every activation is represented once;
we call this assignment \emph{complementary}. Allowing overlap lets the
fine stream encode the coarse stream's residual. This trades coverage for
correction: an overlap is useful when the error removed at one position
exceeds the error introduced by leaving another position uncovered.

\begin{figure}[!tb]
\centering
\includegraphics[width=\columnwidth]{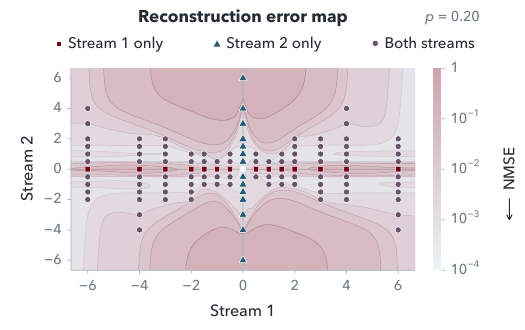}
\vspace{-10pt}
\caption{\textbf{Two streams provide finer reconstruction levels.}
Observed pairs $(a_1/S_1,a_2/S_2)$ from Qwen3.5-4B-Base at $p=0.20$.
Squares and triangles show a contribution from stream 1 or 2 alone;
circles show both. Shading interpolates each pair's mean normalized
squared residual, $((u-a_1-a_2)/S_1)^2$, on a log scale.
Different pairs summarize different inputs.}
\label{fig:stream-interaction}
\end{figure}

Figure~\ref{fig:stream-interaction} shows these choices in
Qwen3.5-4B-Base activations. Squares and triangles on the axes have one
nonzero contribution; off-axis circles have two. The fine contribution can be
positive or negative, adjusting the coarse reconstruction in either
direction. Combining the grids thus makes additional reconstruction
levels available without increasing the FP4 value budget.

Encoding each block requires two choices: the scales of the two grids
and the positions selected by each stream. Jointly searching both online
would repeat assignment for each candidate scale pair. To keep encoding
lightweight, we calibrate the scale ratio offline and fix it at inference.
We then derive both scales from the block maximum and assign values
through two greedy selections.

\subsection{Scale Selection}
\label{sec:compact-scales}
The coarse scale follows the current block maximum, while the calibrated
ratio $p$ determines the fine scale:
\begin{align}
 S_1&=\operatorname{RN}_{\mathrm{UE5M3}}(\max|u|/6),\nonumber\\
 S_2&=\operatorname{RN}_{\mathrm{UE5M3}}(pS_1),\qquad 0<p\leq1.
 \label{eq:compact-scales}
\end{align}
Here $6$ is the largest E2M1 magnitude. For finite, nonnegative inputs,
$\operatorname{RN}_{\mathrm{UE5M3}}$ applies round-to-nearest, ties-to-even
conversion with finite saturation, clamps the scale code to $[1,254]$,
and returns its decoded value. This encoder policy keeps both scales
positive, including for all-zero blocks. Both scales apply to the same
32 positions and adapt to each block, while $p$ stays fixed.

Reducing $p$ gives the fine grid smaller reconstruction steps near zero
but narrows its range. Calibration minimizes activation-induced output
error (Section~\ref{sec:calibration}), leaving no runtime scale search.

\subsection{Greedy Stream Assignment}
\label{sec:assignment}
With the scales fixed, the remaining online decision is which positions
fill each stream's two slots. To avoid searching pairs of sparsity patterns, we
first choose the best complementary assignment, then fix its coarse
stream and select fine contributions from the residual. Both decisions
reduce to selecting two gains within each quartet.

Let
\begin{equation}
 Q(v,S)=S\,\operatorname{RN}_{\mathrm{E2M1}}(v/S)
 \label{eq:quantizer}
\end{equation}
denote quantization followed by dequantization. We compute
$q_1=Q(u,S_1)$ and $q_2=Q(u,S_2)$ for the block. Assigning position
$i$ to the coarse grid instead of the fine grid reduces squared error by
\begin{equation}
 g_{1,i}=(u_i-q_{2,i})^2-(u_i-q_{1,i})^2.
 \label{eq:first-gain}
\end{equation}
Retaining the two largest-gain entries of $q_1$ per quartet and zeroing
the others forms $a_1$. Since complementary error changes only through
these gains, this selection minimizes it at the given scales.

We then quantize the residual $r=u-a_1$. At positions selected by
the first stream, $r$ contains quantization error; elsewhere, it contains
the original value. For $c=Q(r,S_2)$, the benefit of retaining a
fine-stream contribution is
\begin{equation}
 g_{2,i}=r_i^2-(r_i-c_i)^2.
 \label{eq:second-gain}
\end{equation}
Selecting the two largest reductions per quartet forms $a_2$ from the
quantized residual $c$. It can therefore represent an omitted value or
correct the coarse approximation.

The second selection is optimal conditional on $a_1$ and can recover
the complementary mask. At fixed scales in exact arithmetic, the
result therefore has no greater activation reconstruction error than
the best complementary assignment. This local guarantee does not
extend to output error or perplexity.

\subsection{Smoothing}
\label{sec:smoothing}
Large activation channels can determine the scale of an entire block,
reducing the precision available to the remaining values. We use
SmoothQuant~\cite{smoothquant2023} with a default migration strength
$\alpha=0.5$ to redistribute channel magnitude between activations and
weights before quantization. For a positive channel multiplier $b$, we form
\begin{align}
 U&=\operatorname{BF16}(X\odot b),\label{eq:smoothed-input}\\
 W_4&=Q_{\mathrm{FP4,G32,UE5M3}}
 \!\left(W/b\right).
 \label{eq:transformed-weights}
\end{align}
Inverse scaling of the weights preserves the linear transformation
before rounding and quantization. We quantize the transformed weights
with RTN by default.

\subsection{Calibrating the Scale Ratio}
\label{sec:calibration}
Calibration chooses $p$ by layer-output error, accounting for how the
fixed weights amplify activation errors. For projections $j$ sharing a
raw input, we test $\mathcal P=\{0.05,0.10,\ldots,1.00\}$, holding $b$
and $W_4$ fixed. Each candidate reruns scale rounding and both selections
on the calibration inputs. We select
\begin{align}
 p^*&=\underset{p\in\mathcal P}{\operatorname{argmin}}
 \sum_j\|\Delta U^{(j)}(p)(W_4^{(j)})^\top\|_F^2,\nonumber\\
 \Delta U^{(j)}(p)&=A_1^{(j)}(p)+A_2^{(j)}(p)-U^{(j)}.
 \label{eq:calibration}
\end{align}
Here $U^{(j)}$ is projection $j$'s prepared BF16 input. The reference
uses these activations with the same quantized weights, so calibration
targets the error introduced by activation encoding.

After calibration, each new input is encoded into two streams, giving
the approximate layer output
\begin{equation}
 \widehat Y=\operatorname{BF16}\!\left[
 A_1W_4^\top+A_2W_4^\top\right].
 \label{eq:sparsequant}
\end{equation}
Inverse smoothing is absorbed into $W_4$, so the output needs no rescaling.

\section{Execution on Vera Rubin}
\label{sec:system}
Adaptive assignment adds online work and metadata traffic. Efficient
execution must control these costs while processing both streams without
duplicating global-memory weight-tile loads or outputs. We prepare and pack weights offline,
with channel transforms and scale ratios fixed before inference. At runtime,
a fused preparation kernel encodes the activations, followed by a
shared-weight sparse GEMM.

\subsection{Fused Activation Preparation}
Each new input requires selecting and packing two sparsity patterns.
We fuse the channel transform, scale computation, greedy assignment
(Section~\ref{sec:assignment}), and packing into one kernel. The coarse
scale uses the transformed input block's maximum; the fine scale is
derived using the calibrated ratio, avoiding a second maximum reduction
over the residual. The kernel writes FP4 codes, UE5M3 scales, and position
metadata while preserving the original
$K$ positions, so changing sparsity patterns requires no weight rearrangement.

The FP4 value budget is unchanged, but metadata and scales add traffic:
each G32 block stores 16 bytes of values, 8 bytes of position metadata,
and 2 bytes of scales. This is 6.5 bits per logical activation before
padding and replication, compared with 4.5 bits for dense G16 FP4,
excluding any tensor-level factor.

\subsection{Shared-Weight Sparse GEMM}
Both streams multiply the same weights. Our GEMM shares each weight tile
staged in shared memory across both sparse streams, avoiding a separate
global-memory weight-tile load for the second stream. Both products update
one FP32 accumulator. The epilogue converts the result to BF16 without
materializing intermediate products.

Together, the products perform dense GEMM's $2MNK$ nonzero FLOPs.
Our latency measurements include both the activation-preparation kernel
and the shared-weight GEMM, capturing the cost of assignment and additional metadata traffic
(Section~\ref{sec:headline-protocol}).

For a standard projection, preparation reads the incoming activation
tensor. In grouped MoE, preparation for the first expert GEMM follows
token dispatch; preparation for the second follows the intermediate
activation. Routing and the weighted combination of expert outputs
remain unchanged.

\section{Evaluation}
\label{sec:evaluation}

\subsection{Experimental Setup}
\paragraph{Models.}
We begin by evaluating quantization from BF16 on one dense model,
Qwen3.5-4B~\cite{qwen354b2026}, and one MoE model,
Nemotron-3-Nano-30B-A3B~\cite{nemotron3nano2025}. We then study how
\sq{} adapts to natively quantized checkpoints using
GPT-OSS-20B~\cite{gptoss2025},
MiniMax-M3.1-Flash~\cite{minimaxm31config2026}, and
DeepSeek-V4.1-Flash-1017~\cite{deepseekv41flash2026}.
To test its extension to joint language and vision quantization, we
evaluate Qwen3-VL-32B-Instruct~\cite{qwen3vl2025} and
Nemotron-Nano-12B-v2-VL~\cite{nemotronnanovl2025}. We use base checkpoints
for perplexity and post-trained checkpoints for task accuracy.

\paragraph{Quantization.}
Unless specified otherwise, four-bit operands use VRFP4: E2M1 values
with one UE5M3 scale per block. Dense W4A4 uses G16 weights and
activations; \sq{} uses G32 weights and 32 logical positions per
activation stream. W4A16 uses G32 weights and BF16 activations.
We test the composability of \sq{} with vanilla RTN, AWQ, and Four Over
Six~\cite{awq2024,fouroversix2025} to evaluate its applicability across
quantization methods. Appendix~C provides more details on which model components are quantized
and which retain reference precision.

\subsection{Language-Model Quality and Kernel Efficiency}
\label{sec:llm-eval}
% Language-model quality is a standalone float; scaling has its own figure below.
% Sources: tables/data/qwen_quality.json, qwen_reasoning.json, and nemotron_quality.json.
% Latest scores and protocol: tables/data/results_refresh_20261029.json.
% Additional Qwen SQ reasoning results: tables/data/qwen_sq_variants_20261029.json.
% Latest completed cells: tables/data/language_quality_refresh_aaaabdca_20261029.json.
% Four Over Six SQ Base PPL: tables/data/qwen_fos_sq_ppl_20261029.json.
% Earlier displayed-row selection: tables/data/llm_quality_refresh.json.
% Qwen AWQ + SQ CSR170K: tables/data/qwen_awq_sq_csr_20261030.json.
% Refreshed Nemotron reasoning/CSR: tables/data/nemotron_refresh_f2a5c1b9_20261030.json.
% Completed SQ MMLU-Pro/CSR: tables/data/sq_completed_scores_20261030.json.
% Qwen Four Over Six + SQ CSR170K: tables/data/qwen_fos_sq_csr_20261030.json.
% Nemotron AWQ + SQ CSR170K: tables/data/nemotron_awq_sq_csr_20261030.json.
% Exact Nemotron SQ CSR counts and FoS MMLU-Pro: tables/data/nemotron_sq_csr_breakdown_20261030.json.
% Latest completed cells: tables/data/accuracy_missing_cells_20261030.json.
% Latest accuracy.md import: tables/data/table2_accuracy_refresh_0b11db40_20261030.json.
% Nemotron AWQ + SQ GPQA-D: tables/data/table2_accuracy_refresh_3fa08273_20261030.json.
% Final Nemotron Four Over Six + SQ GPQA-D: tables/data/nemotron_fos_gpqa_author_update_20261030.json.
\begin{table*}[!t]
\centering
\caption{\textbf{\sq{} composes with RTN, AWQ, and Four Over Six.}
It lowers perplexity relative to dense W4A4 on both dense and MoE models.
PPL uses base checkpoints; accuracy (\%) uses post-trained versions.
Model-name rows are BF16; W4A16 retains BF16 activations.
All FP4 results use VRFP4 with UE5M3 block scales.}
\label{tab:llm}
\small
\setlength{\tabcolsep}{3pt}
\def\sqrecipebreak{\sqtablebreak{\textwidth}}
% Ordinary fixed-width cells keep the highlight continuous across each row.
\begin{tabular}{
>{\raggedright\arraybackslash}p{\dimexpr.31\textwidth-2\tabcolsep\relax}
>{\centering\arraybackslash}p{\dimexpr.065\textwidth-2\tabcolsep\relax}
>{\centering\arraybackslash}p{\dimexpr.125\textwidth-2\tabcolsep\relax}
>{\centering\arraybackslash}p{\dimexpr.115\textwidth-2\tabcolsep\relax}
*{2}{>{\centering\arraybackslash}p{\dimexpr.13\textwidth-2\tabcolsep\relax}}
>{\centering\arraybackslash}p{\dimexpr.125\textwidth-2\tabcolsep\relax}}
\toprule
\textbf{Method} & \textbf{Group}
& \shortstack{\textbf{WikiText-2}\\PPL$\downarrow$}
& \shortstack{\textbf{CSR}\\6-task Avg.$\uparrow$}
& \textbf{AIME2026}$\uparrow$
& \textbf{GPQA-D}$\uparrow$
& \textbf{MMLU-Pro}$\uparrow$ \\
\midrule
\textbf{Qwen3.5-4B} & \textemdash & 8.05 & 70.2 & 87.8 & 77.1 & 79.5 \\
\sqrecipebreak
\textcolor{gray}{\hspace{0.7em}+ RTN-W (W4A16)} & \textcolor{gray}{32} & \textcolor{gray}{8.43} & \textcolor{gray}{69.1} & \textcolor{gray}{71.1} & \textcolor{gray}{76.8} & \textcolor{gray}{77.3} \\
\hspace{0.7em}+ RTN & 16 & 8.67 & 66.6 & 58.9 & 72.6 & 76.6 \\
\rowcolor{sqshade}
\hspace{1.4em}\textbf{+ \sqshort} & \textbf{32} & \textbf{8.53} & \textbf{68.2} & \textbf{67.8} & \textbf{74.6} & \textbf{76.9} \\
\sqrecipebreak
\textcolor{gray}{\hspace{0.7em}+ AWQ (W4A16)} & \textcolor{gray}{32} & \textcolor{gray}{8.34} & \textcolor{gray}{69.5} & \textcolor{gray}{76.7} & \textcolor{gray}{70.7} & \textcolor{gray}{77.1} \\
\hspace{0.7em}+ AWQ & 16 & 8.57 & 67.9 & 80.0 & 69.9 & 77.2 \\
\rowcolor{sqshade}
\hspace{1.4em}\textbf{+ \sqshort} & \textbf{32} & \textbf{8.48} & \textbf{68.5} & \textbf{77.8} & \textbf{71.7} & \textbf{77.1} \\
\sqrecipebreak
\textcolor{gray}{\hspace{0.7em}+ Four Over Six (W4A16)} & \textcolor{gray}{32} & \textcolor{gray}{8.39} & \textcolor{gray}{69.3} & \textcolor{gray}{74.4} & \textcolor{gray}{73.9} & \textcolor{gray}{77.1} \\
\hspace{0.7em}+ Four Over Six & 16 & 8.57 & 68.4 & 64.4 & 70.7 & 75.3 \\
\rowcolor{sqshade}
\hspace{1.4em}\textbf{+ \sqshort{}} & \textbf{32} & \textbf{8.52} & \textbf{68.6} & \textbf{65.6} & \textbf{69.7} & \textbf{77.0} \\
\midrule
\textbf{Nemotron-3-Nano-30B-A3B} & \textemdash & 5.33 & 69.8 & 87.8 & 59.9 & 72.6 \\
\sqrecipebreak
\textcolor{gray}{\hspace{0.7em}+ RTN-W (W4A16)} & \textcolor{gray}{32} & \textcolor{gray}{6.05} & \textcolor{gray}{69.3} & \textcolor{gray}{86.7} & \textcolor{gray}{60.8} & \textcolor{gray}{68.8} \\
\hspace{0.7em}+ RTN & 16 & 6.85 & 68.5 & 78.9 & 52.7 & 61.6 \\
\rowcolor{sqshade}
\hspace{1.4em}\textbf{+ \sqshort} & \textbf{32} & \textbf{5.57} & \textbf{69.4} & \textbf{86.7} & \textbf{58.8} & \textbf{70.5} \\
\sqrecipebreak
\textcolor{gray}{\hspace{0.7em}+ AWQ (W4A16)} & \textcolor{gray}{32} & \textcolor{gray}{5.48} & \textcolor{gray}{69.4} & \textcolor{gray}{81.1} & \textcolor{gray}{62.6} & \textcolor{gray}{74.9} \\
\hspace{0.7em}+ AWQ & 16 & 5.62 & 68.5 & 81.1 & 61.3 & 71.3 \\
\rowcolor{sqshade}
\hspace{1.4em}\textbf{+ \sqshort} & \textbf{32} & \textbf{5.52} & \textbf{68.9} & \textbf{82.2} & \textbf{60.8} & \textbf{72.8} \\
\sqrecipebreak
\textcolor{gray}{\hspace{0.7em}+ Four Over Six (W4A16)} & \textcolor{gray}{32} & \textcolor{gray}{5.97} & \textcolor{gray}{69.1} & \textcolor{gray}{78.9} & \textcolor{gray}{54.7} & \textcolor{gray}{67.4} \\
\hspace{0.7em}+ Four Over Six & 16 & 6.82 & 69.0 & 80.0 & 52.9 & 64.9 \\
\rowcolor{sqshade}
\hspace{1.4em}\textbf{+ \sqshort} & \textbf{32} & \textbf{5.73} & \textbf{69.4} & \textbf{83.3} & \textbf{62.8} & \textbf{69.7} \\
\bottomrule
\end{tabular}
\end{table*}

\paragraph{Two-stream recipes improve FP4 quality.}
Table~\ref{tab:llm} tests full-model quality across dense and MoE
architectures using WikiText-2 perplexity~\cite{wikitext2016}, a six-task
commonsense reasoning (CSR) average, AIME2026~\cite{aime2026}, GPQA
Diamond~\cite{gpqa2023}, and MMLU-Pro~\cite{mmlupro2024}.
Commonsense is the unweighted zero-shot mean of PIQA~\cite{piqa2020},
ARC-e and ARC-c~\cite{arc2018}, HellaSwag~\cite{hellaswag2019},
WinoGrande~\cite{winogrande2020}, and BoolQ~\cite{boolq2019}, a six-task
subset of the LLM-Adapters benchmarks~\cite{llmadapters2023}.
AIME2026 uses a pinned community transcription of the 30 AIME I/II
questions~\cite{aime2026data}; AIME2026 and GPQA-D average three runs.
MMLU-Pro builds on MMLU~\cite{mmlu2021}; we use one full evaluation pass.

\sq{} composes with RTN, AWQ, and Four Over Six, lowering perplexity
relative to dense W4A4 and recovering much of the corresponding W4A16
performance. On Nemotron-3-Nano-30B-A3B, the RTN and Four Over Six
recipes even achieve lower perplexity than their W4A16 references.

\paragraph{RTN recipe recovers reasoning accuracy.}
On Qwen3.5-4B, \sq{} raises AIME2026 accuracy from 58.9\% to 67.8\%
and GPQA-D from 72.6\% to 74.6\% relative to dense RTN W4A4.
On Nemotron, \sq{} raises AIME2026 accuracy from 78.9\% to
86.7\%, matching the W4A16 reference, and GPQA-D from 52.7\% to 58.8\%.

\paragraph{Native kernels accelerate GEMM and MoE over BF16.}
\label{sec:headline-protocol}
% Author-supplied kernel performance figure, imported 30 October 2026.
% Share the result-table float queue to preserve reading order.
\begin{table*}[!t]
\centering
\includegraphics[width=\textwidth]{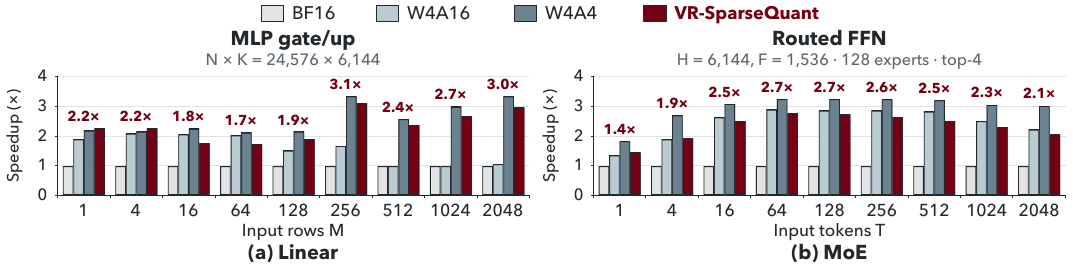}
\vspace{-10pt}
\begingroup
\expandafter\def\csname @captype\endcsname{figure}
\caption{\textbf{Operator speedup and the cost of sparse encoding on Rubin.}
BF16 median latency divided by method median latency for
(a) an MLP gate/up GEMM at TP1 and (b) a routed FFN at TP2, using
synthetic inputs with MiniMax shapes. Timings include online activation
preparation and exclude weight preparation and communication; MoE routes
are precomputed and balanced. W4A4 uses the fastest applicable kernel for
each shape. GEMM uses a PyTorch BF16 baseline; MoE uses TRT-LLM BF16.}
\label{fig:llm-speed}
\endgroup
\end{table*}

Figure~\ref{fig:llm-speed} compares BF16, W4A16, dense W4A4, and our
sparse W4A4 on an MLP gate/up GEMM (TP1) and a routed MoE FFN (TP2).
For dense W4A4, we select the fastest applicable state-of-the-art kernel
evaluated per shape, including FlashInfer's
\texttt{mm\_fp4}~\cite{flashinfer2025,flashinferfp4}
and the \texttt{flashinfer\_cutedsl} and \texttt{trtllm\_moe}
backends~\cite{flashinfercutedslmoe,flashinfermoe}.
Across input sizes from 1 to 2,048, \sq{}
achieves $1.7$--$3.1\times$ speedup over BF16 for GEMM and
$1.4$--$2.7\times$ for MoE. Relative to these optimized W4A4 baselines, \sq{}
has 7--27\% higher latency for the tested GEMMs with $M\geq16$ and
18--47\% higher latency for the tested MoE sizes. Timings include online activation preparation but exclude weight
preparation and communication; the MoE benchmark uses precomputed
balanced routing.

\subsection{Quantizing Full and Linear Attention}
\label{sec:quantization-coverage}
% Sources: tables/data/qwen_coverage.json and coverage_support_refresh_20261029.json.
\begin{table}[!tb]
\centering
\caption{\textbf{\sq{} reduces the perplexity penalty of quantizing attention.}
Cumulative coverage on Qwen3.5-4B-Base extends from MLPs to full
attention and Gated DeltaNet linear attention.}
\label{tab:quantization-coverage}
\sqablationstyle
\begin{tabular}{
>{\raggedright\arraybackslash}p{\dimexpr.24\linewidth-2\tabcolsep\relax}
>{\centering\arraybackslash}p{\dimexpr.12\linewidth-2\tabcolsep\relax}
>{\centering\arraybackslash}p{\dimexpr.20\linewidth-2\tabcolsep\relax}
>{\columncolor{sqshade}\centering\arraybackslash}p{\dimexpr.24\linewidth-2\tabcolsep\relax}
>{\centering\arraybackslash}p{\dimexpr.20\linewidth-2\tabcolsep\relax}}
\toprule
\textbf{Coverage} & \textbf{\# Proj.}
& \shortstack{\textbf{RTN}\\W4A4 G16}
& \shortstack{\textbf{\sqshort{}}\\W4A4 G32}
& \shortstack{\textbf{RTN-W}\\W4A16 G32} \\
\midrule
MLP & 96 & 8.39 & \textbf{8.33} & 8.25 \\
\sqtablebreak{\linewidth}
+ Full attn. & 128 & 8.50 & \textbf{8.39} & 8.30 \\
\sqtablebreak{\linewidth}
+ Linear attn. & 248 & 8.67 & \textbf{8.53} & 8.43 \\
\bottomrule
\end{tabular}
\end{table}

Attention projections can be sensitive to quantization. We test whether
\sq{} improves FP4 quality in both full-attention and linear-attention
layers, extending its benefit beyond MLPs.
Table~\ref{tab:quantization-coverage} starts with 96 MLP projections in
Qwen3.5-4B-Base, then adds 32 full-attention and 120 Gated DeltaNet
projections. Each recipe retains its fitted weights as coverage expands;
excluded projections remain BF16. With full attention included, \sq{}
lowers perplexity from 8.50 to 8.39 relative to dense RTN W4A4; including
linear attention lowers it from 8.67 to 8.53. The smaller perplexity
increase at each step shows that \sq{} reduces the quality penalty of
quantizing both attention families.

\subsection{Applying \sq{} to Pre-Quantized Models}
\label{sec:agentic-eval}
\label{sec:quantized-adaptation}
We also evaluate \sq{} on models released with quantized weights, including
GPT-OSS-20B, MiniMax-M3.1-Flash, and DeepSeek-V4.1-Flash-1017.
Their native differences in group size and scale format provide a broad
range of settings for testing \sq{}.

\paragraph{GPT-OSS-20B.}
% Source: tables/data/gpt_oss_native_mxfp4.json (pinned receipts and author update).
% SQ results: tables/data/gpt_oss_sq_refresh_pr65_20261029.json.
\begin{table}[!tb]
\centering
\caption{\textbf{\sq{} improves GPT-OSS-20B GPQA-D accuracy over dense FP4
under both scale formats.} Decoded expert weights remain unchanged.}
\label{tab:native-mxfp4}
\sqablationstyle
\begin{tabular}{
>{\raggedright\arraybackslash}m{\dimexpr.32\linewidth-2\tabcolsep\relax}
>{\centering\arraybackslash}m{\dimexpr.18\linewidth-2\tabcolsep\relax}
>{\centering\arraybackslash}m{\dimexpr.18\linewidth-2\tabcolsep\relax}
>{\centering\arraybackslash}m{\dimexpr.08\linewidth-2\tabcolsep\relax}
>{\centering\arraybackslash}m{\dimexpr.24\linewidth-2\tabcolsep\relax}}
\toprule
\multirow{2}{*}{\textbf{Method}}
& \multicolumn{3}{c}{\textbf{Scaling Type / Group Size}}
& \multirow{2}{*}{\textbf{GPQA-D}$\uparrow$} \\
\cmidrule(lr){2-4}
& W & A & G & \\
\midrule
\textcolor{gray}{W4A16} & \textcolor{gray}{UE8M0}
& \textcolor{gray}{\textemdash} & \textcolor{gray}{32}
& \textcolor{gray}{71.7} \\
\sqtablebreak{\linewidth}
W4A4 & UE8M0 & UE8M0 & 32 & 69.4 \\
\rowcolor{sqshade}
\hspace{0.5em}\textbf{+ \sqshort{}} & UE8M0 & UE8M0 & 32 & \textbf{70.9} \\
\sqtablebreak{\linewidth}
W4A4 & UE5M3 & UE5M3 & 16 & 69.5 \\
\rowcolor{sqshade}
\hspace{0.5em}\textbf{+ \sqshort{}} & UE5M3 & UE5M3 & 32 & \textbf{70.7} \\
\bottomrule
\end{tabular}
\end{table}

GPT-OSS-20B stores its expert weights in MXFP4 with G32, matching the
weight-group size required by our sparse Tensor Core path. We can
therefore apply \sq{} without requantizing the weights. The first W4A4
pair in Table~\ref{tab:native-mxfp4} retains the native UE8M0 scales and
shows improved GPQA-D accuracy with \sq{}. All weight scales used by
this checkpoint are also exactly representable in UE5M3, allowing us to
recode them into VRFP4 without changing the E2M1 codes or decoded weights.
UE5M3 also permits finer activation-scale choices than power-of-two
UE8M0 scaling. In both formats, \sq{} recovers much of the GPQA-D accuracy
lost when activations are quantized to FP4. With UE8M0 scales, it reaches
70.9\% accuracy, compared with 71.7\% for W4A16.

% Completed scores: tables/data/frontier_quality.json (author update, 29 October 2026).
% Earlier MiniMax SQ results: tables/data/frontier_sq_refresh_cdc4da9d_20261029.json.
% Latest author results: tables/data/frontier_author_update_20261030.json.
% DeepSeek all-layer completion: tables/data/deepseek_all_sq_author_update_20261030.json.
\begin{table*}[!t]
\centering
\caption{\textbf{Applying \sq{} to natively quantized models.}
Relative to the native checkpoints, expert-only MiniMax improves on three
of four tasks, and all-linear DeepSeek improves BFCL and $\tau^2$-Bench.
W/A denote weight/activation formats;
G is group size. VRFP4 uses UE5M3 block scales; NVFP4 combines FP8 block
and FP32 global scales. Scores are percentages.
$^{\dagger}$~DeepSeek's native FP8 linear weights use $32\times32$ tiles;
activations use G32.}
\label{tab:frontier}
\footnotesize
\setlength{\tabcolsep}{1.5pt}
\renewcommand{\arraystretch}{1.0}
\begin{tabular}{
>{\raggedright\arraybackslash}p{\dimexpr0.154\textwidth-2\tabcolsep\relax}
>{\centering\arraybackslash}p{\dimexpr0.070\textwidth-2\tabcolsep\relax}
>{\centering\arraybackslash}p{\dimexpr0.070\textwidth-2\tabcolsep\relax}
>{\centering\arraybackslash}p{\dimexpr0.033\textwidth-2\tabcolsep\relax}
>{\centering\arraybackslash}p{\dimexpr0.070\textwidth-2\tabcolsep\relax}
>{\centering\arraybackslash}p{\dimexpr0.070\textwidth-2\tabcolsep\relax}
>{\centering\arraybackslash}p{\dimexpr0.033\textwidth-2\tabcolsep\relax}
>{\centering\arraybackslash}p{\dimexpr0.125\textwidth-2\tabcolsep\relax}
>{\centering\arraybackslash}p{\dimexpr0.125\textwidth-2\tabcolsep\relax}
>{\centering\arraybackslash}p{\dimexpr0.125\textwidth-2\tabcolsep\relax}
>{\centering\arraybackslash}p{\dimexpr0.125\textwidth-2\tabcolsep\relax}}
\toprule
\multirow{2.5}{*}{\textbf{Method}}
& \multicolumn{3}{c}{\textbf{Experts}}
& \multicolumn{3}{c}{\textbf{Linears}}
& \multirow{2.5}{*}{\shortstack{\textbf{BFCL v4}\\\textbf{Avg.}$\uparrow$}}
& \multirow{2.5}{*}{\shortstack{\textbf{$\tau^2$-Bench}\\\textbf{Avg.}$\uparrow$}}
& \multirow{2.5}{*}{\shortstack{\textbf{Automation}\\\textbf{Bench}$\uparrow$}}
& \multirow{2.5}{*}{\shortstack{\textbf{CharXiv}\\\textbf{RQ}$\uparrow$}} \\
\cmidrule(lr){2-4}\cmidrule(lr){5-7}
& W & A & G & W & A & G & & & & \\
\midrule
\rowcolor{boxfill}
\multicolumn{11}{c}{\rule{0pt}{2.2ex}\textbf{MiniMax-M3.1-Flash}} \\
\addlinespace[1.5pt]
Native & NVFP4 & NVFP4 & 16 & MXFP8 & MXFP8 & 32 & 64.3 & 84.0 & 61.5 & 82.3 \\
\rowcolor{sqshade}
\hspace{0.5em}\textbf{+ \sqshort} & VRFP4 & VRFP4 & 32 & MXFP8 & MXFP8 & 32 & \textbf{64.9} & \textbf{84.6} & \textbf{59.2} & \textbf{82.6} \\
\sqtablebreak{\textwidth}
VRFP4 (all) & VRFP4 & VRFP4 & 16 & VRFP4 & VRFP4 & 16 & 65.0 & 83.9 & 62.5 & 81.5 \\
\rowcolor{sqshade}
\hspace{0.5em}\textbf{+ \sqshort} & VRFP4 & VRFP4 & 32 & VRFP4 & VRFP4 & 32 & \textbf{64.5} & \textbf{83.7} & \textbf{62.7} & \textbf{81.5} \\
\midrule
\rowcolor{boxfill}
\multicolumn{11}{c}{\rule{0pt}{2.2ex}\textbf{DeepSeek-V4.1-Flash-1017}} \\
\addlinespace[1.5pt]
Native & MXFP4 & MXFP8 & 32 & FP8 & MXFP8 & 32$^{\dagger}$ & 73.0 & 88.2 & 60.3 & 77.9 \\
MXFP4 (experts) & MXFP4 & MXFP4 & 32 & FP8 & MXFP8 & 32$^{\dagger}$ & 73.6 & 87.9 & 58.7 & 77.0 \\
\rowcolor{sqshade}
\hspace{0.5em}\textbf{+ \sqshort} & VRFP4 & VRFP4 & 32 & FP8 & MXFP8 & 32$^{\dagger}$ & \textbf{73.9} & \textbf{87.7} & \textbf{59.0} & \textbf{77.7} \\
\sqtablebreak{\textwidth}
VRFP4 (linears) & MXFP4 & MXFP4 & 32 & VRFP4 & VRFP4 & 16 & 73.9 & 89.1 & 58.8 & 75.6 \\
\rowcolor{sqshade}
\hspace{0.5em}\textbf{+ \sqshort} & VRFP4 & VRFP4 & 32 & VRFP4 & VRFP4 & 32 & \textbf{74.3} & \textbf{90.4} & \textbf{59.2} & \textbf{75.5} \\
\bottomrule
\end{tabular}
\end{table*}

For MiniMax and DeepSeek, we apply \sq{} first to the experts alone and
then to all eligible linear projections (Table~\ref{tab:frontier}).
We evaluate BFCL v4~\cite{bfclv4,bfclv4protocol2025},
$\tau^2$-Bench~\cite{tau2bench2025},
AutomationBench~\cite{automationbench2026}, and CharXiv reasoning
questions (RQ)~\cite{charxiv2024}. BFCL scores average four blocks;
$\tau^2$-Bench scores average domain pass$^1$. We use GPT-5.5 to judge
CharXiv and maximum reasoning effort for all runs, keeping evaluation
settings fixed within each model.

\paragraph{MiniMax-M3.1-Flash.}
MiniMax uses NVFP4 G16 for its experts and MXFP8 for other linear
projections~\cite{minimaxm31config2026,minimaxm31native2026}.
For \sq{}, we requantize the decoded expert weights to VRFP4 G32
with RTN. This conversion merges G16 groups and folds NVFP4's global
factor into UE5M3 block scales, adding some weight-quantization error.
The expert-only variant outperforms the native checkpoint on BFCL and
CharXiv, achieving the highest CharXiv score among the tested MiniMax
variants (Table~\ref{tab:frontier}).
When applied to all eligible linear projections, \sq{} raises
AutomationBench from 61.5\% to 62.7\%.

% These are quality comparisons: \sq{} uses four
% Vera Rubin GPUs, while Native uses B300.

\paragraph{DeepSeek-V4.1-Flash-1017.}
DeepSeek's native checkpoint uses MXFP4 G32 expert weights and MXFP8 G32
activations~\cite{deepseekv41flash2026,nvidiamxfp8}.
On the expert projections, \sq{} improves three of four tasks over
dense MXFP4 activation quantization. Applying \sq{} to all eligible
linear projections brings both weights and activations to VRFP4 and
reaches \textbf{74.3\% BFCL} and \textbf{90.4\% $\tau^2$-Bench},
compared with 73.0\% and 88.2\% for the native checkpoint.
Quantizing the MoE activations to FP4 also moves these projections from
the native W4A8 format to W4A4, allowing our kernels to use Rubin's
higher-throughput compact FP4 Tensor Core
path~\cite{nvidiaptx2026,nvidiarubin}.
This setting includes shared-expert, attention, indexer, and dense
feed-forward projections; other operators, caches, and states retain
reference precision.

\subsection{Vision-Language Quality}
\label{sec:vlm-eval}
% Sources: tables/data/qwen3vl_quality.json and nemotron_vl_quality.json.
% Qwen SQ results: tables/data/results_refresh_20261029.json.
% Nemotron SQ interim results: tables/data/assignment_vlm_refresh_ba17785d_20261029.json.
% Nemotron SQ completion: tables/data/nemotron_vl_completion_20261030.json.
\begin{table*}[!t]
\centering
\caption{\textbf{FP4 quantization of both language and vision backbones.}
\sq{} improves the mean over dense RTN W4A4 and stays within 0.7 points
of TinyChat while using native FP4 Tensor Core execution.
DocVQA uses ANLS. $\dagger$~Our TinyChat implementation combines dynamic
FP8 vision quantization with AWQ-style W4A16 language quantization.}
\label{tab:vlm}
\small
\setlength{\tabcolsep}{2pt}
\def\sqbaselinebreak{\sqtablebreak{\textwidth}}
\begin{tabular}{
>{\raggedright\arraybackslash}p{\dimexpr.27\textwidth-2\tabcolsep\relax}
*{2}{>{\centering\arraybackslash}p{\dimexpr.145\textwidth-2\tabcolsep\relax}}
*{4}{>{\centering\arraybackslash}p{\dimexpr.09\textwidth-2\tabcolsep\relax}}
>{\centering\arraybackslash}p{\dimexpr.08\textwidth-2\tabcolsep\relax}}
\toprule
\multirow{2}{*}{\textbf{Method}} & \multicolumn{2}{c}{\textbf{Precision}}
& \multirow{2}{*}{\textbf{MMMU}} & \multirow{2}{*}{\textbf{DocVQA}}
& \multirow{2}{*}{\textbf{TextVQA}} & \multirow{2}{*}{\textbf{ChartQA}}
& \multirow{2}{*}{\textbf{Avg.}$\uparrow$} \\
\cmidrule(lr){2-3}
& Language model & Vision encoder & & & & & \\
\midrule
Qwen3-VL-32B-Instruct & BF16 & BF16 & 62.7 & 96.1 & 83.5 & 87.1 & 82.4 \\
\sqbaselinebreak
\hspace{0.7em}+ RTN & W4A4 & W4A4 & 61.8 & 95.6 & 81.1 & 86.3 & 81.2 \\
\hspace{0.7em}+ TinyChat$^{\dagger}$ & W4A16 & W8A8 & 63.2 & 95.9 & 83.0 & 86.7 & 82.2 \\
\rowcolor{sqshade}
\hspace{0.7em}\textbf{+ \sqshort} & W4A4 & W4A4 & \textbf{62.8} & \textbf{96.1} & \textbf{82.1} & \textbf{86.1} & \textbf{81.8} \\
\midrule
Nemotron-Nano-12B-v2-VL & BF16 & BF16 & 64.6 & 94.3 & 82.8 & 89.6 & 82.8 \\
\sqbaselinebreak
\hspace{0.7em}+ RTN & W4A4 & W4A4 & 63.1 & 93.4 & 82.0 & 89.0 & 81.9 \\
\hspace{0.7em}+ TinyChat$^{\dagger}$ & W4A16 & W8A8 & 65.6 & 94.1 & 82.8 & 89.2 & 82.9 \\
\rowcolor{sqshade}
\hspace{0.7em}\textbf{+ \sqshort} & W4A4 & W4A4 & \textbf{63.3} & \textbf{94.1} & \textbf{82.0} & \textbf{89.6} & \textbf{82.2} \\
\bottomrule
\end{tabular}
\end{table*}

Joint backbone quantization tests the representation across language
and visual activation distributions. We evaluate
MMMU~\cite{mmmu2024},
DocVQA~\cite{docvqa2021}, TextVQA~\cite{textvqa2019}, and
ChartQA~\cite{chartqa2022}
and report their unweighted mean, using ANLS for DocVQA. The multimodal
connector and visual-token settings remain fixed. Our TinyChat baseline
combines AWQ W4A16 G16 language weights with dynamic E4M3 W8A8 vision
quantization~\cite{awq2024,tinychat2024}.

We combine channel smoothing with AWQ G32 weights in both backbones.
\sq{} improves the average score over dense RTN W4A4 on both models
(Table~\ref{tab:vlm}), while executing the quantized language and vision
projections directly on Rubin's FP4 Tensor Cores.

\subsection{Ablation Studies}
\label{sec:analysis}
These studies separate activation encoding from smoothing and examine
scale ratios, sparsity patterns, and weight-group size. Unless stated
otherwise, quality studies use
Qwen3.5-4B-Base WikiText-2 perplexity with fixed calibration data,
projections, and evaluation tokens. Hyperparameters are selected on
calibration data alone.

% RTN values: tables/data/scale_format.json, direct UE5M3, all 248 projections.
% RTN-W G32: tables/data/weight_group_sensitivity.json.
% SQ values: tables/data/results_refresh_20261029.json.
\begin{table}[!htb]
\centering
\caption{\textbf{Two-stream encoding lowers perplexity before smoothing.}
The smoothed recipe reduces it further.}
\label{tab:ablation-components}
\sqablationstyle
\newcommand{\sqenabled}{\textcolor{green!45!black}{{\footnotesize$\checkmark$}}}
\newcommand{\sqdisabled}{\textcolor{red!75!black}{{\footnotesize$\times$}}}
\begin{tabular}{
>{\raggedright\arraybackslash}m{\dimexpr.33\linewidth-2\tabcolsep\relax}
>{\centering\arraybackslash}m{\dimexpr.13\linewidth-2\tabcolsep\relax}
>{\centering\arraybackslash}m{\dimexpr.20\linewidth-2\tabcolsep\relax}
>{\centering\arraybackslash}m{\dimexpr.15\linewidth-2\tabcolsep\relax}
>{\centering\arraybackslash}m{\dimexpr.19\linewidth-2\tabcolsep\relax}}
\toprule
\textbf{Method} & \textbf{Group} & \textbf{Smoothing}
& \textbf{Sparsity}
& \shortstack{\textbf{WikiText-2}\\PPL$\downarrow$} \\
\midrule
RTN-W (W4A16) & 32 & \pending & \pending & 8.43 \\
RTN W4A4 & 32 & \pending & \pending & 8.86 \\
RTN W4A4 & 16 & \pending & \pending & 8.67 \\
\sqtablebreak{\linewidth}
\rowcolor{sqshade}
\sqshort{} W4A4 & 32 & \sqdisabled & \sqenabled & 8.58 \\
\rowcolor{sqshade}
\sqshort{} W4A4 & 32 & \sqenabled & \sqenabled & 8.53 \\
\bottomrule
\end{tabular}
\tabnote{Group size applies to quantized operands. Smoothing uses
$\alpha=0.5$, rebuilt RTN weights, and recalibrated $p$. Dashes in the
component columns mean not applied.}
\end{table}

\paragraph{Encoding improves quality before smoothing.}
With fixed RTN G32 weights, no smoothing, and 248 projections,
Table~\ref{tab:ablation-components} isolates activation encoding:
replacing dense FP4 activations with two streams lowers perplexity from
8.86 to 8.58. Rebuilding RTN weights and recalibrating $p$ after smoothing,
under the same budget, reduces it further to 8.53.

% Author-supplied PDF, copied unchanged on 22 October 2026.
% Measured curve data and source hashes: figures/data/p_sweep.json.
\begin{figure}[!htb]
\centering
\includegraphics[width=\columnwidth]{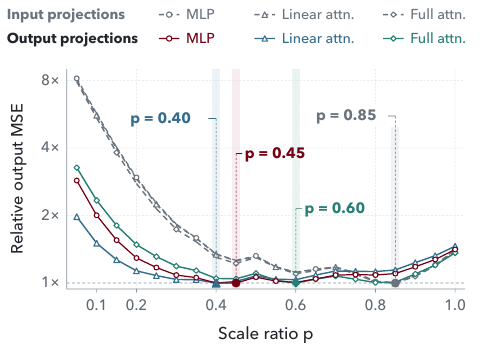}
\caption{\textbf{Different projections favor different scale ratios.}
Scale-ratio study of a weighted-assignment variant on Qwen3.5-4B.
Output MSE is pooled by projection type and normalized by its minimum
over 20 ratios (log axis). Dashed/solid curves denote input/output
projections; filled markers identify minima. These curves motivate
separate calibration for different projections.}
\label{fig:p-sweep}
\end{figure}

\paragraph{Different projections favor different scale ratios.}
The fine scale balances resolution of small residuals against coverage
of larger values. For the weighted-assignment variant in
Figure~\ref{fig:p-sweep}, input projections favor $p=0.85$, while output
projections favor $0.40$--$0.60$. The different minima motivate selecting
a ratio for each shared-input group through the output-error objective
in Section~\ref{sec:calibration}.

% Source: tables/data/assignment_vlm_refresh_ba17785d_20261029.json.
\begin{table}[!htb]
\centering
\definecolor{sqmaskcoarse}{HTML}{850026}
\definecolor{sqmaskfine}{HTML}{32738E}
\caption{\textbf{Assignment and scale calibration.}
\sq{} gives lower perplexity than complementary splitting with the same
smoothed RTN G32 weights. Complementary uses independent stream scales; \sq{} uses
calibrated $p$.
Schematic masks show \textcolor{sqmaskcoarse}{coarse} above
\textcolor{sqmaskfine}{fine}.}
\label{tab:ablation-assignment}
\vspace{3pt}
\sqablationstyle
\renewcommand{\arraystretch}{1.0}
\definecolor{sqmaskedge}{HTML}{D4DDE4}
\newcommand{\sqmaskcell}[1]{\begingroup
  \setlength{\fboxsep}{0pt}\setlength{\fboxrule}{0.3pt}%
  \fcolorbox{sqmaskedge}{#1}{\rule{0pt}{6.5pt}\rule{6.5pt}{0pt}}%
  \endgroup\hspace{0.8pt}}
\newcommand{\sqmaskempty}{\sqmaskcell{white}}
\newcommand{\sqmaskpair}[2]{\raisebox{0pt}%
  [\height][\depth]%
  {\shortstack{#1\\[-1pt]#2}}}
% Illustrative sparsity patterns: disjoint streams versus overlapping streams.
\begin{tabular}{
>{\raggedright\arraybackslash}m{\dimexpr.46\linewidth-2\tabcolsep\relax}
>{\centering\arraybackslash}m{\dimexpr.30\linewidth-2\tabcolsep\relax}
>{\centering\arraybackslash}m{\dimexpr.24\linewidth-2\tabcolsep\relax}}
\toprule
\textbf{Assignment}
& \textbf{Masks}
& \shortstack{\textbf{WikiText-2}\\PPL$\downarrow$} \\
\midrule
Complementary & \sqmaskpair
  {\sqmaskcell{sqmaskcoarse}\sqmaskcell{sqmaskcoarse}\sqmaskempty\sqmaskempty}
  {\sqmaskempty\sqmaskempty\sqmaskcell{sqmaskfine}\sqmaskcell{sqmaskfine}}
& 8.73 \\
\sqtablebreak{\linewidth}
\rowcolor{sqshade}
\textbf{\sqshort{}} & \sqmaskpair
  {\sqmaskcell{sqmaskcoarse}\sqmaskempty\sqmaskcell{sqmaskcoarse}\sqmaskempty}
  {\sqmaskcell{sqmaskfine}\sqmaskcell{sqmaskfine}\sqmaskempty\sqmaskempty}
& \textbf{8.53} \\
\bottomrule
\end{tabular}
\end{table}

\begin{figure}[!htb]
\centering
\includegraphics[width=\columnwidth]{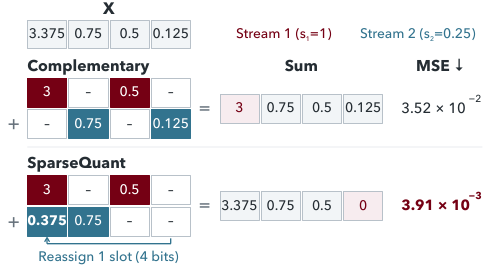}
\caption{\textbf{Trading coverage for correction.}
In this constructed quartet, reallocating one fine-stream slot to a
residual reduces reconstruction MSE by $9\times$.
Both encodings retain two values per stream at the same scales, with
\sq{}'s coarse selection fixed. This comparison isolates overlap;
Table~\ref{tab:ablation-assignment} also changes the scale rule.}
\label{fig:assignment}
\end{figure}

\paragraph{Residual correction competes with complete coverage.}
Table~\ref{tab:ablation-assignment} compares complete coverage with
residual correction using identical smoothed RTN G32 weights.
Complementary splitting assigns the two largest magnitudes per quartet
to one stream and the remaining values to the other, with independent
scales. It gives 8.73 PPL, versus 8.53 for \sq{}'s calibrated ratio and
error-based assignment (Eqs.~\ref{eq:first-gain}--\ref{eq:second-gain}).
This gain measures assignment and scaling together, not overlap alone.

Figure~\ref{fig:assignment} shows the mechanism at fixed scales:
moving one fine-stream slot from a small value to a residual correction
reduces the constructed quartet's MSE by $9\times$, with the coarse
stream unchanged. Correcting one position can therefore remove more
error than omitting another introduces.

% Source: tables/data/coverage_support_refresh_20261029.json.
% Dense G32 RTN control: tables/data/scale_format.json; author-confirmed 8.86.
\begin{table}[!htb]
\centering
\caption{\textbf{Elementwise 2:4 sparsity lowers perplexity relative to
paired 4:8 and approaches flexible 16:32.} Simulated Qwen3.5-4B-Base
comparisons with shared G32 VRFP4 weights, smoothing, and calibrated $p$
across sparse variants.}
\label{tab:ablation-representation}
\sqablationstyle
\begin{tabular}{
>{\raggedright\arraybackslash}m{\dimexpr.46\linewidth-2\tabcolsep\relax}
>{\raggedright\arraybackslash}m{\dimexpr.34\linewidth-2\tabcolsep\relax}
>{\centering\arraybackslash}m{\dimexpr.20\linewidth-2\tabcolsep\relax}}
\toprule
\textbf{Structure} & \textbf{Selection per stream}
& \shortstack{\textbf{WikiText-2}\\PPL$\downarrow$} \\
\midrule
Dense Quantization\,{\scriptsize\textcolor{draftgray}{(Control)}}
& All 32 values & 8.86 \\
\sqtablebreak{\linewidth}
Paired 4:8\,{\scriptsize\textcolor{draftgray}{(Blackwell)}}
& 2 of 4 adjacent pairs & 8.61 \\
\sqtablebreak{\linewidth}
\rowcolor{sqshade}
\textbf{Elementwise 2:4}\,{\scriptsize\textcolor{draftgray}{(Vera Rubin)}}
& Any 2 of 4 values & \textbf{8.53} \\
\sqtablebreak{\linewidth}
Flexible 16:32\,{\scriptsize\textcolor{draftgray}{(Oracle)}}
& Any 16 of 32 values & 8.49 \\
\bottomrule
\end{tabular}
\end{table}

\paragraph{Elementwise structured sparsity improves representation quality.}
Paired 4:8 couples neighboring values; Vera Rubin's elementwise 2:4 selects
individual values per quartet (Figure~\ref{fig:gpu-capabilities}).
Table~\ref{tab:ablation-representation} tests simulated quality with
fitted G32 weights, smoothing, and $p$ fixed across sparse variants.
Elementwise selection lowers perplexity from 8.61 to 8.53; allowing any
16 positions per block with the same sequential greedy rule, rather than
exhaustive search, reaches 8.49. Thus, 2:4 captures most of the improvement
over paired selection, with a further gain from relaxation. Neither
hardware pattern family contains the other; both are subsets of 16:32.

% Full-precision source and computed relative changes: tables/data/weight_group_sensitivity.json.
\begin{table}[!htb]
\centering
\caption{\textbf{Weight quantization at G16 and G32.}
Increasing the weight block size from 16 to 32 incurs a smaller perplexity
penalty with AWQ than with RTN.}
\label{tab:ablation-weights}
\small
\setlength{\tabcolsep}{3pt}
\begin{tabular*}{\columnwidth}{@{\extracolsep{\fill}}lrrr@{}}
\toprule
\textbf{Quantizer} & \shortstack{\textbf{G16}\\PPL$\downarrow$}
& \shortstack{\textbf{G32}\\PPL$\downarrow$}
& \shortstack{\textbf{Relative}\\\textbf{increase}$\downarrow$} \\
\midrule
RTN & 8.37 & 8.43 & +0.76\% \\
\sqtablebreak{\columnwidth}
AWQ & 8.31 & 8.34 & +0.33\% \\
\bottomrule
\end{tabular*}
\end{table}

\paragraph{AWQ reduces the weight-group penalty.}
To measure the cost of the sparse path's G32 weights,
Table~\ref{tab:ablation-weights} retains BF16 activations and varies
weight-group size within each quantizer. Increasing G16 to G32 raises perplexity by
0.76\% for RTN and 0.33\% for AWQ. Both quantize the same 248
projections with E2M1/UE5M3 and no additional smoothing; AWQ retains its
channel scaling~\cite{awq2024}. AWQ reduces this weight-side cost,
complementing the improvements in activation encoding.

\section{Related Work}
\label{sec:related}
\paragraph{Dense low-bit quantization.}
SmoothQuant transfers activation dynamic range to weights, while AWQ
protects sensitive weight channels using activation
statistics~\cite{smoothquant2023,awq2024}. QuaRot and SpinQuant use
rotations, and FlatQuant learns affine transformations, to improve
dense weight and activation quantization~\cite{quarot2024,spinquant2025,flatquant2025}.
For microscaled FP4, Four Over Six selects between two block scales,
and MR-GPTQ combines format-specific quantization with blockwise
rotations~\cite{fouroversix2025,mrgptq2025}. These methods improve dense
quantizers' coordinates or grids. \sq{} instead allocates
FP4 contributions across positions through sparse metadata and remains
complementary to weight quantization and channel smoothing.

\paragraph{Structured sparse decomposition.}
TASD approximates weights and activations with multiple structured sparse
tensors and sums their products~\cite{tasd2025}. SDQ decomposes weights
into structured sparse components with different precisions and
error-aware selection~\cite{sdq2024}. \sq{}'s two FP4 activation components
can overlap, letting one value correct another's rounding error. Their
combined value count equals dense FP4's, so correction competes with
representing distinct positions within a fixed allocation.

\paragraph{Residual correction for quantization.}
SVDQuant absorbs outliers into a high-precision low-rank branch and
quantizes the residual to four bits~\cite{svdquant2025}. ARCQuant adds
FP4 activation-residual channels to the reduction dimension of a single
GEMM~\cite{arcquant2026}. SharQ uses a sparse FP4 backbone and a dense
FP4 residual, sharing one weight payload across the two
products~\cite{sharq2026}. Rollout-ResQ adds a 2:4 FP4 residual to dense
FP4 activations for reinforcement-learning rollouts~\cite{rolloutresq2026}.
\sq{} constrains both components to 2:4 sparsity. Its residual stream
must choose between covering omitted values and correcting represented
ones; a calibrated scale ratio and online error-based selection make
this choice without searching scale pairs at inference.

\paragraph{Sparse execution.}
Haziza et al.\ exploit Squared-ReLU activation sparsity with online
2:4 packing and fused FP8 quantization~\cite{activation24sparsity2025}.
Attention and token sparsity operate at a different level, reducing
the tokens or spatial locations processed~\cite{duoattention2024,lserve2025,sparserefine2024,sparsevila2025}.
For low-bit execution, TinyChat, QServe, and Nunchaku control
dequantization, data movement, and branch
overhead~\cite{awq2024,qserve2024,svdquant2025}. ARCQuant and SharQ
integrate residual preparation with FP4 execution~\cite{arcquant2026,sharq2026}.
\sq{} reuses weight tiles and an accumulator for both sparse products
in one GEMM on Rubin's compact elementwise FP4 path~\cite{nvidiaptx2026}. SpAtten~\cite{wang2021spatten} proposes KV Cache sparsification and quantization, we well as the combinations of them.

\section{Limitations}
\label{sec:limitations}
\sq{} improves activation precision through additional encoding work and
metadata traffic, which can increase latency relative to dense W4A4.
The sparse FP4 Tensor Core path we use requires G32 weights, sacrificing
scaling granularity relative to the dense G16 baseline. This can introduce
additional weight quantization error that activation encoding cannot
directly correct. Stronger weight quantizers are therefore particularly
important for mitigating this penalty and retaining the gains from
activation encoding (Table~\ref{tab:ablation-weights}).
Reported speedups cover GEMM and MoE operators, excluding communication
and other serving overheads. Native kernel measurements use Vera Rubin;
the sparsity-pattern comparisons assess simulated quality. Results for
complete recipes and converted checkpoints reflect both weight and
activation effects, while the fixed-weight ablations isolate activation
encoding.

\section{Conclusion}
\label{sec:conclusion}
To the best of our knowledge, this is the first study to design and
evaluate adaptive activation quantization using Vera Rubin's compact
elementwise 2:4 sparse FP4 instructions. \sq{} uses sparsity patterns to
allocate precision: two overlapping streams refine selected activations
while retaining dense FP4's value count and nonzero arithmetic. Offline
calibration, greedy assignment, and a shared-weight GEMM make this
allocation practical. Our kernels use Vera Rubin's direct UE5M3 block
scaling for E2M1 FP4, which we term VRFP4.

Across RTN, AWQ, and Four Over Six, \sq{} improves perplexity relative
to dense W4A4, with further applications to vision-language models and
released low-bit checkpoints. Including online activation preparation,
native Rubin kernels achieve up to $3.1\times$ GEMM and $2.7\times$ MoE
speedup over BF16.

More broadly, Rubin's elementwise selection and VRFP4 scaling show how
new hardware capabilities expand the design space of quantization.
More flexible sparsity patterns, lower metadata costs, and tighter fusion
of encoding with surrounding operators could make adaptive precision
allocation more efficient. These results motivate developing quantization
formats, algorithms, and kernels together as GPU hardware evolves, with
model quality, memory traffic, and latency as joint design objectives.

\bibliography{main.bib}

@inproceedings{awq2024,
  title = {{AWQ}: Activation-aware Weight Quantization for On-Device {LLM} Compression and Acceleration},
  author = {Ji Lin and Jiaming Tang and Haotian Tang and Shang Yang and Wei-Ming Chen and Wei-Chen Wang and Guangxuan Xiao and Xingyu Dang and Chuang Gan and Song Han},
  booktitle = {Proceedings of Machine Learning and Systems},
  volume = {6},
  year = {2024},
  url = {https://arxiv.org/abs/2306.00978},
  eprint = {2306.00978},
  archivePrefix = {arXiv}
}

@inproceedings{smoothquant2023,
  title = {{SmoothQuant}: Accurate and Efficient Post-Training Quantization for Large Language Models},
  author = {Guangxuan Xiao and Ji Lin and Mickael Seznec and Hao Wu and Julien Demouth and Song Han},
  booktitle = {Proceedings of the 40th International Conference on Machine Learning},
  series = {Proceedings of Machine Learning Research},
  volume = {202},
  pages = {38087--38099},
  publisher = {PMLR},
  year = {2023},
  url = {https://proceedings.mlr.press/v202/xiao23c.html},
  eprint = {2211.10438},
  archivePrefix = {arXiv}
}

@inproceedings{qserve2024,
  title = {{QServe}: {W4A8KV4} Quantization and System Co-design for Efficient {LLM} Serving},
  author = {Yujun Lin and Haotian Tang and Shang Yang and Zhekai Zhang and Guangxuan Xiao and Chuang Gan and Song Han},
  booktitle = {Proceedings of Machine Learning and Systems},
  volume = {7},
  year = {2025},
  url = {https://arxiv.org/abs/2405.04532},
  eprint = {2405.04532},
  archivePrefix = {arXiv}
}

@inproceedings{lserve2025,
  title = {{LServe}: Efficient Long-sequence {LLM} Serving with Unified Sparse Attention},
  author = {Shang Yang and Junxian Guo and Haotian Tang and Qinghao Hu and Guangxuan Xiao and Jiaming Tang and Yujun Lin and Zhijian Liu and Yao Lu and Song Han},
  booktitle = {Proceedings of Machine Learning and Systems},
  volume = {7},
  year = {2025},
  url = {https://arxiv.org/abs/2502.14866},
  eprint = {2502.14866},
  archivePrefix = {arXiv}
}

@inproceedings{sparserefine2024,
  title = {Sparse Refinement for Efficient High-Resolution Semantic Segmentation},
  author = {Zhijian Liu and Zhuoyang Zhang and Samir Khaki and Shang Yang and Haotian Tang and Chenfeng Xu and Kurt Keutzer and Song Han},
  booktitle = {European Conference on Computer Vision},
  year = {2024},
  url = {https://arxiv.org/abs/2407.19014},
  eprint = {2407.19014},
  archivePrefix = {arXiv}
}

@article{sparsevila2025,
  title = {{SparseVILA}: Decoupling Visual Sparsity for Efficient {VLM} Inference},
  author = {Samir Khaki and Junxian Guo and Jiaming Tang and Shang Yang and Yukang Chen and Konstantinos N. Plataniotis and Yao Lu and Song Han and Zhijian Liu},
  journal = {arXiv preprint arXiv:2510.17777},
  year = {2025},
  url = {https://arxiv.org/abs/2510.17777},
  eprint = {2510.17777},
  archivePrefix = {arXiv}
}

@article{fouroversix2025,
  title = {Four Over Six: More Accurate {NVFP4} Quantization with Adaptive Block Scaling},
  author = {Jack Cook and Junxian Guo and Guangxuan Xiao and Yujun Lin and Keith Wyss and Mahdi Nazemi and Asit Mishra and Carlo del Mundo and Tijmen Blankevoort and Song Han},
  journal = {arXiv preprint arXiv:2512.02010},
  year = {2025},
  url = {https://arxiv.org/abs/2512.02010},
  eprint = {2512.02010},
  archivePrefix = {arXiv}
}

@article{sharq2026,
  title = {{SharQ}: Bridging Activation Sparsity and {FP4} Quantization for {LLM} Inference},
  author = {Haoqian Meng and Yilun Luo and Yafei Zhao and Wenyuan Liu and Huaqing Zheng and Xindian Ma and Peng Zhang},
  journal = {arXiv preprint arXiv:2606.26587},
  year = {2026},
  url = {https://arxiv.org/abs/2606.26587},
  eprint = {2606.26587},
  archivePrefix = {arXiv}
}

@manual{nvidiaptx2026,
  title = {Parallel Thread Execution {ISA}, Version 9.4},
  author = {{NVIDIA}},
  organization = {NVIDIA},
  year = {2026},
  url = {https://docs.nvidia.com/cuda/parallel-thread-execution/index.html},
  note = {Accessed October 5, 2026}
}

@manual{nvidiaptx87,
  title = {Parallel Thread Execution {ISA}, Version 8.7},
  author = {{NVIDIA}},
  year = {2025},
  url = {https://docs.nvidia.com/cuda/archive/12.8.1/parallel-thread-execution/index.html},
  note = {Sparse block scaling and FP4 metadata; accessed October 20, 2026}
}

@manual{nvidiaa100,
  title = {{NVIDIA A100 Tensor Core GPU} Datasheet},
  author = {{NVIDIA}},
  year = {2026},
  url = {https://images.nvidia.com/data-center/a100/a100-datasheet.pdf},
  note = {80GB SXM specifications; accessed October 20, 2026}
}

@manual{nvidiah100,
  title = {{NVIDIA H100 Tensor Core GPU} Specifications},
  author = {{NVIDIA}},
  year = {2026},
  url = {https://www.nvidia.com/en-us/data-center/h100/},
  note = {SXM specifications; accessed October 20, 2026}
}

@manual{nvidiablackwell,
  title = {{NVIDIA Blackwell Architecture} Datasheet},
  author = {{NVIDIA}},
  year = {2026},
  url = {https://dam-cdn.nvd.orangelogic.com/AssetLink/y441155802qub41q118b2852i557jem5.pdf},
  note = {Individual HGX B200 GPU specifications, p. 8; accessed October 20, 2026}
}

@manual{nvidiarubin,
  title = {{NVIDIA HGX}: Individual {Rubin GPU} Specifications},
  author = {{NVIDIA}},
  year = {2026},
  url = {https://www.nvidia.com/en-us/data-center/hgx/},
  note = {Sparse NVFP4 inference specification; accessed October 20, 2026}
}

@manual{nvidiarubinplatform2026,
  title = {{NVIDIA Vera Rubin Platform}},
  author = {{NVIDIA}},
  year = {2026},
  url = {https://www.nvidia.com/en-us/data-center/technologies/rubin/},
  note = {Accessed October 29, 2026}
}

@manual{nvidianvfp4,
  title = {{NVFP4}: Two-Level Scaling},
  author = {{NVIDIA}},
  organization = {{NVIDIA Transformer Engine}},
  year = {2026},
  url = {https://nvidia.github.io/TransformerEngine/features/low_precision_training/nvfp4/nvfp4.html},
  note = {Accessed October 5, 2026}
}

@misc{nemotronnanovl2025,
  title = {{NVIDIA Nemotron Nano 12B V2 VL BF16}: Model Card},
  author = {{NVIDIA}},
  year = {2025},
  howpublished = {Hugging Face model repository},
  url = {https://huggingface.co/nvidia/NVIDIA-Nemotron-Nano-12B-v2-VL-BF16},
  note = {Accessed October 26, 2026}
}

@misc{qwen354b2026,
  title = {{Qwen3.5-4B}: Model Card},
  author = {{Qwen Team}},
  year = {2026},
  howpublished = {Hugging Face model repository},
  version = {851bf6e806efd8d0a36b00ddf55e13ccb7b8cd0a},
  url = {https://huggingface.co/Qwen/Qwen3.5-4B},
  note = {Accessed September 30, 2026}
}

@misc{qwen354bbase2026,
  title = {{Qwen3.5-4B-Base}: Model Card},
  author = {{Qwen Team}},
  year = {2026},
  howpublished = {Hugging Face model repository},
  url = {https://huggingface.co/Qwen/Qwen3.5-4B-Base},
  note = {Accessed October 30, 2026}
}

@inproceedings{mmlu2021,
  title = {Measuring Massive Multitask Language Understanding},
  author = {Dan Hendrycks and Collin Burns and Steven Basart and Andy Zou and Mantas Mazeika and Dawn Song and Jacob Steinhardt},
  booktitle = {International Conference on Learning Representations},
  year = {2021},
  url = {https://arxiv.org/abs/2009.03300},
  eprint = {2009.03300},
  archivePrefix = {arXiv}
}

@inproceedings{mmlupro2024,
  title = {{MMLU-Pro}: A More Robust and Challenging Multi-Task Language Understanding Benchmark},
  author = {Yubo Wang and Xueguang Ma and Ge Zhang and Yuansheng Ni and Abhranil Chandra and Shiguang Guo and Weiming Ren and Aaran Arulraj and Xuan He and Ziyan Jiang and Tianle Li and Max Ku and Kai Wang and Alex Zhuang and Rongqi Fan and Xiang Yue and Wenhu Chen},
  booktitle = {Advances in Neural Information Processing Systems},
  year = {2024},
  url = {https://arxiv.org/abs/2406.01574},
  eprint = {2406.01574},
  archivePrefix = {arXiv}
}

@article{gpqa2023,
  title = {{GPQA}: A Graduate-Level Google-Proof {Q\&A} Benchmark},
  author = {David Rein and Betty Li Hou and Asa Cooper Stickland and Jackson Petty and Richard Yuanzhe Pang and Julien Dirani and Julian Michael and Samuel R. Bowman},
  journal = {arXiv preprint arXiv:2311.12022},
  year = {2023},
  url = {https://arxiv.org/abs/2311.12022},
  eprint = {2311.12022},
  archivePrefix = {arXiv}
}

@misc{qwen3vl2025,
  title = {{Qwen3-VL-32B-Instruct}: Model Card},
  author = {{Qwen Team}},
  year = {2025},
  howpublished = {Hugging Face model repository},
  version = {0cfaf48183f594c314753d30a4c4974bc75f3ccb},
  url = {https://huggingface.co/Qwen/Qwen3-VL-32B-Instruct},
  note = {Accessed September 30, 2026}
}

@misc{nemotron3nano2025,
  title = {{NVIDIA-Nemotron-3-Nano-30B-A3B-BF16}: Model Card},
  author = {{NVIDIA}},
  year = {2025},
  howpublished = {Hugging Face model repository},
  url = {https://huggingface.co/nvidia/NVIDIA-Nemotron-3-Nano-30B-A3B-BF16},
  note = {Released December 15, 2025. Accessed September 30, 2026}
}

@misc{minimaxm31config2026,
  title = {{MiniMax-M3.1-Flash}: Released Quantization Configuration},
  author = {{MiniMax}},
  year = {2026},
  howpublished = {Hugging Face model repository},
  url = {https://huggingface.co/MiniMaxAI/MiniMax-M3.1-Flash/blob/main/config.json},
  note = {Accessed October 24, 2026}
}

@misc{minimaxm31native2026,
  title = {{MiniMax-M3.1-Flash}: Native Inference Formats},
  author = {{NVIDIA}},
  year = {2026},
  howpublished = {NVIDIA NIM model card},
  url = {https://build.nvidia.com/minimaxai/minimax-m3.1-flash/modelcard},
  note = {Accessed October 24, 2026}
}

@misc{deepseekv41flash2026,
  title = {{DeepSeek-V4.1-Flash-1017}: Reference Inference Implementation},
  author = {{DeepSeek-AI}},
  year = {2026},
  howpublished = {Hugging Face model repository},
  url = {https://huggingface.co/deepseek-ai/DeepSeek-V4.1-Flash-1017/blob/main/inference/model.py},
  note = {Accessed October 24, 2026}
}

@article{tau2bench2025,
  title = {{$\tau^2$-Bench}: Evaluating Conversational Agents in a Dual-Control Environment},
  author = {Victor Barres and Honghua Dong and Soham Ray and Xujie Si and Karthik Narasimhan},
  journal = {arXiv preprint arXiv:2506.07982},
  year = {2025},
  url = {https://arxiv.org/abs/2506.07982}
}

@inproceedings{llmadapters2023,
  title = {{LLM-Adapters}: An Adapter Family for Parameter-Efficient Fine-Tuning of Large Language Models},
  author = {Zhiqiang Hu and Lei Wang and Yihuai Lan and Wanyu Xu and Ee-Peng Lim and Lidong Bing and Xing Xu and Soujanya Poria and Roy Lee},
  booktitle = {Proceedings of the 2023 Conference on Empirical Methods in Natural Language Processing},
  pages = {5254--5276},
  publisher = {Association for Computational Linguistics},
  year = {2023},
  doi = {10.18653/v1/2023.emnlp-main.319},
  url = {https://aclanthology.org/2023.emnlp-main.319/}
}

@misc{wikitext2016,
  title = {Pointer Sentinel Mixture Models},
  author = {Stephen Merity and Caiming Xiong and James Bradbury and Richard Socher},
  year = {2016},
  eprint = {1609.07843},
  archivePrefix = {arXiv},
  primaryClass = {cs.CL},
  url = {https://arxiv.org/abs/1609.07843}
}

@misc{tinychat2024,
  title = {{TinyChat 2.0}: Accelerating Edge {AI} with Efficient {LLM} and {VLM} Deployment},
  author = {Shang Yang and Yuming Lou and Junxian Guo and Jiaming Tang and Haotian Tang and Song Han},
  year = {2024},
  howpublished = {MIT HAN Lab},
  url = {https://hanlab.mit.edu/blog/tinychat20},
  note = {Published December 12, 2024. Accessed September 30, 2026}
}

@inproceedings{mmmu2024,
  title = {{MMMU}: A Massive Multi-discipline Multimodal Understanding and Reasoning Benchmark for Expert {AGI}},
  author = {Xiang Yue and Yuansheng Ni and Kai Zhang and Tianyu Zheng and Ruoqi Liu and Ge Zhang and Samuel Stevens and Dongfu Jiang and Weiming Ren and Yuxuan Sun and Cong Wei and Botao Yu and Ruibin Yuan and Renliang Sun and Ming Yin and Boyuan Zheng and Zhenzhu Yang and Yibo Liu and Wenhao Huang and Huan Sun and Yu Su and Wenhu Chen},
  booktitle = {Proceedings of the IEEE/CVF Conference on Computer Vision and Pattern Recognition},
  pages = {9556--9567},
  year = {2024},
  url = {https://openaccess.thecvf.com/content/CVPR2024/html/Yue_MMMU_A_Massive_Multi-discipline_Multimodal_Understanding_and_Reasoning_Benchmark_for_CVPR_2024_paper.html},
  eprint = {2311.16502},
  archivePrefix = {arXiv}
}

@inproceedings{textvqa2019,
  title = {Towards {VQA} Models That Can Read},
  author = {Amanpreet Singh and Vivek Natarajan and Meet Shah and Yu Jiang and Xinlei Chen and Dhruv Batra and Devi Parikh and Marcus Rohrbach},
  booktitle = {Proceedings of the IEEE/CVF Conference on Computer Vision and Pattern Recognition},
  pages = {8317--8326},
  year = {2019},
  url = {https://openaccess.thecvf.com/content_CVPR_2019/html/Singh_Towards_VQA_Models_That_Can_Read_CVPR_2019_paper.html},
  eprint = {1904.08920},
  archivePrefix = {arXiv}
}

@inproceedings{chartqa2022,
  title = {{ChartQA}: A Benchmark for Question Answering about Charts with Visual and Logical Reasoning},
  author = {Ahmed Masry and Do Xuan Long and Jia Qing Tan and Shafiq Joty and Enamul Hoque},
  booktitle = {Findings of the Association for Computational Linguistics: ACL 2022},
  pages = {2263--2279},
  publisher = {Association for Computational Linguistics},
  year = {2022},
  doi = {10.18653/v1/2022.findings-acl.177},
  url = {https://aclanthology.org/2022.findings-acl.177/}
}

@inproceedings{bfclv4,
  title = {The Berkeley Function Calling Leaderboard ({BFCL}): From Tool Use to Agentic Evaluation of Large Language Models},
  author = {Shishir G Patil and Huanzhi Mao and Fanjia Yan and Charlie Cheng-Jie Ji and Vishnu Suresh and Ion Stoica and Joseph E. Gonzalez},
  booktitle = {Proceedings of the 42nd International Conference on Machine Learning},
  series = {Proceedings of Machine Learning Research},
  volume = {267},
  pages = {48371--48392},
  publisher = {PMLR},
  year = {2025},
  url = {https://proceedings.mlr.press/v267/patil25a.html}
}

@manual{nvidiamxfp8,
  title = {{MXFP8}},
  author = {{NVIDIA}},
  organization = {NVIDIA},
  year = {2026},
  url = {https://docs.nvidia.com/deeplearning/transformer-engine/features/low_precision_training/mxfp8/mxfp8.html},
  note = {Transformer Engine documentation. Accessed October 24, 2026}
}

@inproceedings{docvqa2021,
  title = {{DocVQA}: A Dataset for VQA on Document Images},
  author = {Mathew, Minesh and Karatzas, Dimosthenis and Jawahar, C. V.},
  booktitle = {Proceedings of the IEEE/CVF Winter Conference on Applications of Computer Vision},
  year = {2021},
  url = {https://arxiv.org/abs/2007.00398}
}

@inproceedings{svdquant2025,
  title = {{SVDQuant}: Absorbing Outliers by Low-Rank Components for 4-Bit Diffusion Models},
  author = {Muyang Li and Yujun Lin and Zhekai Zhang and Tianle Cai and Xiuyu Li and Junxian Guo and Enze Xie and Chenlin Meng and Jun-Yan Zhu and Song Han},
  booktitle = {International Conference on Learning Representations},
  year = {2025},
  url = {https://arxiv.org/abs/2411.05007},
  eprint = {2411.05007},
  archivePrefix = {arXiv}
}

@article{duoattention2024,
  title = {{DuoAttention}: Efficient Long-Context {LLM} Inference with Retrieval and Streaming Heads},
  author = {Guangxuan Xiao and Jiaming Tang and Jingwei Zuo and Junxian Guo and Shang Yang and Haotian Tang and Yao Fu and Song Han},
  journal = {arXiv preprint arXiv:2410.10819},
  year = {2024},
  url = {https://arxiv.org/abs/2410.10819},
  eprint = {2410.10819},
  archivePrefix = {arXiv}
}

@misc{gptoss2025,
  title={{gpt-oss-120b} and {gpt-oss-20b} Model Card},
  author={{OpenAI}},
  year={2025},
  eprint={2508.10925},
  archivePrefix={arXiv},
  primaryClass={cs.CL},
  url={https://arxiv.org/abs/2508.10925}
}

@inproceedings{tasd2025,
  title = {Enabling Unstructured Sparse Acceleration on Structured Sparse Accelerators},
  author = {Geonhwa Jeong and Po-An Tsai and Abhimanyu Rajeshkumar Bambhaniya and Stephen W. Keckler and Tushar Krishna},
  booktitle = {Proceedings of Machine Learning and Systems},
  volume = {7},
  year = {2025},
  url = {https://proceedings.mlsys.org/paper_files/paper/2025/hash/e2ec2530db26b54d0b3b060c1e4a1bda-Abstract-Conference.html},
  eprint = {2403.07953},
  archivePrefix = {arXiv}
}

@inproceedings{arcquant2026,
  title = {{ARCQuant}: Boosting {NVFP4} Quantization with Augmented Residual Channels for {LLMs}},
  author = {Haoqian Meng and Yilun Luo and Yafei Zhao and Wenyuan Liu and Peng Zhang and Xindian Ma},
  booktitle = {Proceedings of the 64th Annual Meeting of the Association for Computational Linguistics (Volume 1: Long Papers)},
  pages = {8609--8623},
  publisher = {Association for Computational Linguistics},
  address = {San Diego, California, United States},
  year = {2026},
  doi = {10.18653/v1/2026.acl-long.388},
  url = {https://aclanthology.org/2026.acl-long.388/},
  eprint = {2601.07475},
  archivePrefix = {arXiv}
}

@article{rolloutresq2026,
  title = {{HiFloat4} Format for End-To-End Reinforcement Learning Post-Training of Large Language Models},
  author = {Hei Yi Mak and Shadan Golestan and Hoang Le and Mehran Taghian Jazi and Yunke Peng and Yaoyuan Wang and Yao Wang and Junsong Wang and Tianchi Hu and Fengchen He and Guipeng Hu and Tanzila Rahman and Anandharaju Durai Raju},
  journal = {arXiv preprint arXiv:2607.26515},
  year = {2026},
  url = {https://arxiv.org/abs/2607.26515},
  eprint = {2607.26515},
  archivePrefix = {arXiv}
}

@inproceedings{mrgptq2025,
  title = {Bridging the Gap Between Promise and Performance for Microscaling {FP4} Quantization},
  author = {Vage Egiazarian and Roberto L. Castro and Denis Kuznedelev and Andrei Panferov and Eldar Kurti{\'c} and Shubhra Pandit and Alexandre Marques and Mark Kurtz and Saleh Ashkboos and Torsten Hoefler and Dan Alistarh},
  booktitle = {International Conference on Learning Representations},
  year = {2026},
  url = {https://proceedings.iclr.cc/paper_files/paper/2026/hash/b87bb4f6346d727b265088235e5bc389-Abstract-Conference.html},
  eprint = {2509.23202},
  archivePrefix = {arXiv}
}

@article{sdq2024,
  title = {{SDQ}: Sparse Decomposed Quantization for {LLM} Inference},
  author = {Geonhwa Jeong and Po-An Tsai and Stephen W. Keckler and Tushar Krishna},
  journal = {arXiv preprint arXiv:2406.13868},
  year = {2024},
  url = {https://arxiv.org/abs/2406.13868},
  eprint = {2406.13868},
  archivePrefix = {arXiv}
}

@inproceedings{activation24sparsity2025,
  title = {Accelerating Transformer Inference and Training with {2:4} Activation Sparsity},
  author = {Daniel Haziza and Timothy Chou and Dhruv Choudhary and Jesse Cai and Luca Wehrstedt and Francisco Massa and Jiecao Yu and Geonhwa Jeong and Supriya Rao and Patrick Labatut},
  booktitle = {ICLR 2025 Workshop on Sparsity in LLMs (SLLM)},
  year = {2025},
  url = {https://openreview.net/pdf?id=O5feVk7p6Y},
  eprint = {2503.16672},
  archivePrefix = {arXiv}
}

@inproceedings{quarot2024,
  title = {{QuaRot}: Outlier-Free 4-Bit Inference in Rotated {LLMs}},
  author = {Saleh Ashkboos and Amirkeivan Mohtashami and Maximilian L. Croci and Bo Li and Pashmina Cameron and Martin Jaggi and Dan Alistarh and Torsten Hoefler and James Hensman},
  booktitle = {Advances in Neural Information Processing Systems},
  volume = {37},
  year = {2024},
  doi = {10.52202/079017-3180},
  url = {https://papers.nips.cc/paper_files/paper/2024/hash/b5b939436789f76f08b9d0da5e81af7c-Abstract-Conference.html},
  eprint = {2404.00456},
  archivePrefix = {arXiv}
}

@inproceedings{spinquant2025,
  title = {{SpinQuant}: {LLM} Quantization with Learned Rotations},
  author = {Zechun Liu and Changsheng Zhao and Igor Fedorov and Bilge Soran and Dhruv Choudhary and Raghuraman Krishnamoorthi and Vikas Chandra and Yuandong Tian and Tijmen Blankevoort},
  booktitle = {International Conference on Learning Representations},
  year = {2025},
  url = {https://proceedings.iclr.cc/paper_files/paper/2025/hash/e5b1c0d4866f72393c522c8a00eed4eb-Abstract-Conference.html},
  eprint = {2405.16406},
  archivePrefix = {arXiv}
}

@inproceedings{flatquant2025,
  title = {{FlatQuant}: Flatness Matters for {LLM} Quantization},
  author = {Yuxuan Sun and Ruikang Liu and Haoli Bai and Han Bao and Kang Zhao and Yuening Li and Jiaxin Hu and Xianzhi Yu and Lu Hou and Chun Yuan and Xin Jiang and Wulong Liu and Jun Yao},
  booktitle = {Proceedings of the 42nd International Conference on Machine Learning},
  series = {Proceedings of Machine Learning Research},
  volume = {267},
  pages = {57587--57613},
  publisher = {PMLR},
  year = {2025},
  url = {https://proceedings.mlr.press/v267/sun25l.html},
  eprint = {2410.09426},
  archivePrefix = {arXiv}
}

@article{microscaling2023,
  title = {Microscaling Data Formats for Deep Learning},
  author = {Darvish Rouhani, Bita and Zhao, Ritchie and More, Ankit and Hall, Mathew and Khodamoradi, Alireza and Deng, Summer and Choudhary, Dhruv and Cornea, Marius and Dellinger, Eric and Denolf, Kristof and Stosic, Dusan and Elango, Venmugil and Golub, Maximilian and Heinecke, Alexander and James-Roxby, Phil and Jani, Dharmesh and Kolhe, Gaurav and Langhammer, Martin and Li, Ada and Melnick, Levi and Mesmakhosroshahi, Maral and Rodriguez, Andres and Schulte, Michael and Shafipour, Rasoul and Shao, Lei and Siu, Michael and Dubey, Pradeep and Micikevicius, Paulius and Naumov, Maxim and Verrilli, Colin and Wittig, Ralph and Burger, Doug and Chung, Eric},
  journal = {arXiv preprint arXiv:2310.10537},
  year = {2023},
  url = {https://arxiv.org/abs/2310.10537},
  eprint = {2310.10537},
  archivePrefix = {arXiv}
}

@misc{aime2026,
  title = {2025--26 {AIME} Thresholds Are Now Available},
  author = {{Mathematical Association of America}},
  year = {2025},
  month = dec,
  howpublished = {Official competition announcement},
  url = {https://maa.org/news/2025-26-aime-thresholds-are-now-available/},
  note = {Published December 15, 2025; announces AIME I and II in February 2026}
}

@misc{aime2026data,
  title = {{aime26}},
  author = {{math-ai}},
  year = {2026},
  howpublished = {Hugging Face dataset},
  url = {https://huggingface.co/datasets/math-ai/aime26/tree/79037aebdb6580008fb960d17cb21fd3099083e3},
  note = {Community transcription of AIME 2026 I and II. Revision 79037aebdb6580008fb960d17cb21fd3099083e3. Accessed October 30, 2026}
}

@inproceedings{piqa2020,
  title = {{PIQA}: Reasoning about Physical Commonsense in Natural Language},
  author = {Yonatan Bisk and Rowan Zellers and Ronan Le Bras and Jianfeng Gao and Yejin Choi},
  booktitle = {Proceedings of the AAAI Conference on Artificial Intelligence},
  year = {2020},
  volume = {34},
  number = {05},
  pages = {7432--7439},
  doi = {10.1609/aaai.v34i05.6239},
  url = {https://ojs.aaai.org/index.php/AAAI/article/view/6239}
}

@article{arc2018,
  title = {Think you have Solved Question Answering? Try {ARC}, the {AI2} Reasoning Challenge},
  author = {Peter Clark and Isaac Cowhey and Oren Etzioni and Tushar Khot and Ashish Sabharwal and Carissa Schoenick and Oyvind Tafjord},
  journal = {arXiv preprint arXiv:1803.05457},
  year = {2018},
  url = {https://arxiv.org/abs/1803.05457},
  eprint = {1803.05457},
  archivePrefix = {arXiv}
}

@inproceedings{hellaswag2019,
  title = {{HellaSwag}: Can a Machine Really Finish Your Sentence?},
  author = {Rowan Zellers and Ari Holtzman and Yonatan Bisk and Ali Farhadi and Yejin Choi},
  booktitle = {Proceedings of the 57th Annual Meeting of the Association for Computational Linguistics},
  year = {2019},
  pages = {4791--4800},
  doi = {10.18653/v1/P19-1472},
  url = {https://aclanthology.org/P19-1472/}
}

@inproceedings{winogrande2020,
  title = {{WinoGrande}: An Adversarial Winograd Schema Challenge at Scale},
  author = {Keisuke Sakaguchi and Ronan Le Bras and Chandra Bhagavatula and Yejin Choi},
  booktitle = {Proceedings of the AAAI Conference on Artificial Intelligence},
  year = {2020},
  volume = {34},
  number = {05},
  pages = {8732--8740},
  doi = {10.1609/aaai.v34i05.6399},
  url = {https://ojs.aaai.org/index.php/AAAI/article/view/6399}
}

@inproceedings{boolq2019,
  title = {{BoolQ}: Exploring the Surprising Difficulty of Natural Yes/No Questions},
  author = {Christopher Clark and Kenton Lee and Ming-Wei Chang and Tom Kwiatkowski and Michael Collins and Kristina Toutanova},
  booktitle = {Proceedings of the 2019 Conference of the North American Chapter of the Association for Computational Linguistics: Human Language Technologies, Volume 1 (Long and Short Papers)},
  year = {2019},
  pages = {2924--2936},
  doi = {10.18653/v1/N19-1300},
  url = {https://aclanthology.org/N19-1300/}
}

@article{automationbench2026,
  title = {{AutomationBench}},
  author = {Daniel Shepard and Robin Salimans},
  journal = {arXiv preprint arXiv:2604.18934},
  year = {2026},
  url = {https://arxiv.org/abs/2604.18934},
  eprint = {2604.18934},
  archivePrefix = {arXiv}
}

@inproceedings{charxiv2024,
  title = {{CharXiv}: Charting Gaps in Realistic Chart Understanding in Multimodal {LLMs}},
  author = {Zirui Wang and Mengzhou Xia and Luxi He and Howard Chen and Yitao Liu and Richard Zhu and Kaiqu Liang and Xindi Wu and Haotian Liu and Sadhika Malladi and Alexis Chevalier and Sanjeev Arora and Danqi Chen},
  booktitle = {Advances in Neural Information Processing Systems},
  volume = {37},
  year = {2024},
  note = {Datasets and Benchmarks Track},
  doi = {10.52202/079017-3609},
  url = {https://proceedings.neurips.cc/paper_files/paper/2024/hash/cdf6f8e9fd9aeaf79b6024caec24f15b-Abstract-Datasets_and_Benchmarks_Track.html},
  eprint = {2406.18521},
  archivePrefix = {arXiv}
}

@misc{bfclv4protocol2025,
  title = {{BFCL V4}: Web Search},
  author = {Huanzhi Mao and Raymond Tsao and Jingzhuo Zhou and Shishir G. Patil and Joseph E. Gonzalez},
  year = {2025},
  month = jul,
  howpublished = {Official Berkeley Function Calling Leaderboard documentation},
  url = {https://gorilla.cs.berkeley.edu/blogs/15_bfcl_v4_web_search.html},
  note = {Released July 17, 2025; updated July 20, 2025. Includes V4 score composition}
}

@inproceedings{flashinfer2025,
  title = {{FlashInfer}: Efficient and Customizable Attention Engine for {LLM} Inference Serving},
  author = {Zihao Ye and Lequn Chen and Ruihang Lai and Wuwei Lin and Yineng Zhang and Stephanie Wang and Tianqi Chen and Baris Kasikci and Vinod Grover and Arvind Krishnamurthy and Luis Ceze},
  booktitle = {Proceedings of Machine Learning and Systems},
  volume = {7},
  year = {2025},
  url = {https://proceedings.mlsys.org/paper_files/paper/2025/hash/dbf02b21d77409a2db30e56866a8ab3a-Abstract-Conference.html},
  eprint = {2501.01005},
  archivePrefix = {arXiv}
}

@manual{flashinferfp4,
  title = {{flashinfer.\allowbreak{}gemm.\allowbreak{}mm\_\allowbreak{}fp4}},
  author = {{FlashInfer Contributors}},
  organization = {FlashInfer},
  version = {0.7.2.post1},
  url = {https://docs.flashinfer.ai/generated/flashinfer.gemm.mm_fp4.html},
  note = {FlashInfer 0.7.2.post1 API documentation. Accessed October 30, 2026}
}

@manual{flashinfermoe,
  title = {{flashinfer.\allowbreak{}fused\_\allowbreak{}moe.\allowbreak{}trtllm\_\allowbreak{}fp4\_\allowbreak{}block\_\allowbreak{}scale\_\allowbreak{}moe}},
  author = {{FlashInfer Contributors}},
  organization = {FlashInfer},
  version = {0.7.2.post1},
  url = {https://docs.flashinfer.ai/generated/flashinfer.fused_moe.trtllm_fp4_block_scale_moe.html},
  note = {FlashInfer 0.7.2.post1 API documentation. Accessed October 30, 2026}
}

@manual{flashinfercutedslmoe,
  title = {{flashinfer.\allowbreak{}fused\_\allowbreak{}moe.\allowbreak{}cute\_\allowbreak{}dsl\_\allowbreak{}fused\_\allowbreak{}moe\_\allowbreak{}nvfp4}},
  author = {{FlashInfer Contributors}},
  organization = {FlashInfer},
  version = {0.7.2.post1},
  url = {https://docs.flashinfer.ai/generated/flashinfer.fused_moe.cute_dsl_fused_moe_nvfp4.html},
  note = {FlashInfer 0.7.2.post1 API documentation. Accessed October 30, 2026}
}

@manual{nvidiafp8blockwise,
  title = {{FP8} Blockwise Scaling},
  author = {{NVIDIA}},
  organization = {NVIDIA Transformer Engine},
  version = {2.22.0.dev0},
  url = {https://nvidia.github.io/TransformerEngine/features/low_precision_training/fp8_blockwise_scaling/fp8_blockwise_scaling.html},
  note = {Transformer Engine 2.22.0.dev0 documentation. Accessed October 30, 2026}
}

@misc{kda2026,
  title = {{Kernel Design Agents (KDA)}},
  author = {{NVlabs}},
  year = {2026},
  howpublished = {GitHub repository},
  url = {https://github.com/NVlabs/kda},
  note = {Accessed October 30, 2026}
}

@inproceedings{wang2021spatten,
  title={Spatten: Efficient sparse attention architecture with cascade token and head pruning},
  author={Wang, Hanrui and Zhang, Zhekai and Han, Song},
  booktitle={2021 IEEE international symposium on high-performance computer architecture (HPCA)},
  pages={97--110},
  year={2021},
  organization={IEEE}
}
\bibliographystyle{mlsys2025}
\clearpage
\section*{AI Usage Disclosure}

Generative AI tools, primarily OpenAI Codex and Claude Code, were used to
assist with code generation, writing, and initial figure designs. For code
generation, these tools primarily supported the development and iterative
refinement of computational kernels. Verifiable objectives guided this
process, enabling successive improvements in kernel efficiency through an
approach similar to that explored in Kernel Design Agents
(KDA)~\citep{kda2026}.

AI tools were also used to refine the manuscript's wording and improve the
clarity and readability of explanations, helping communicate the work more
effectively.

\end{document}